\documentclass[lettersize,journal]{IEEEtran}
\usepackage{amsmath,amsfonts}
\usepackage{algorithmic}
\usepackage{algorithm}
\usepackage{array}
\usepackage[caption=false,font=normalsize,labelfont=sf,textfont=sf]{subfig}
\usepackage{textcomp}
\usepackage{stfloats}
\usepackage{url}
\usepackage{verbatim}
\usepackage{graphicx}
\usepackage{cite}
\usepackage[colorlinks,linkcolor=blue]{hyperref} 
\usepackage{mathrsfs}
\usepackage[flushleft]{threeparttable}
\usepackage{tablefootnote}
\usepackage{multirow}
\usepackage{graphicx}
\usepackage{color}
\usepackage{tikz}
\usetikzlibrary{calc,arrows,decorations.markings}
\usepackage{balance}
\usepackage{booktabs}
\usepackage{xpatch}
\usepackage{mathrsfs}
\usepackage{float}

\makeatletter
\usepackage{etoolbox,lipsum}
\patchcmd\@makecaption{\\}{.~}{}{\fail}

\usepackage{caption}

\newcommand{\insertfig}{

  \setcounter{figure}{0} % 重置图形计数器
  \vspace*{0.2in}
  \centering
  \includegraphics[width=\textwidth]{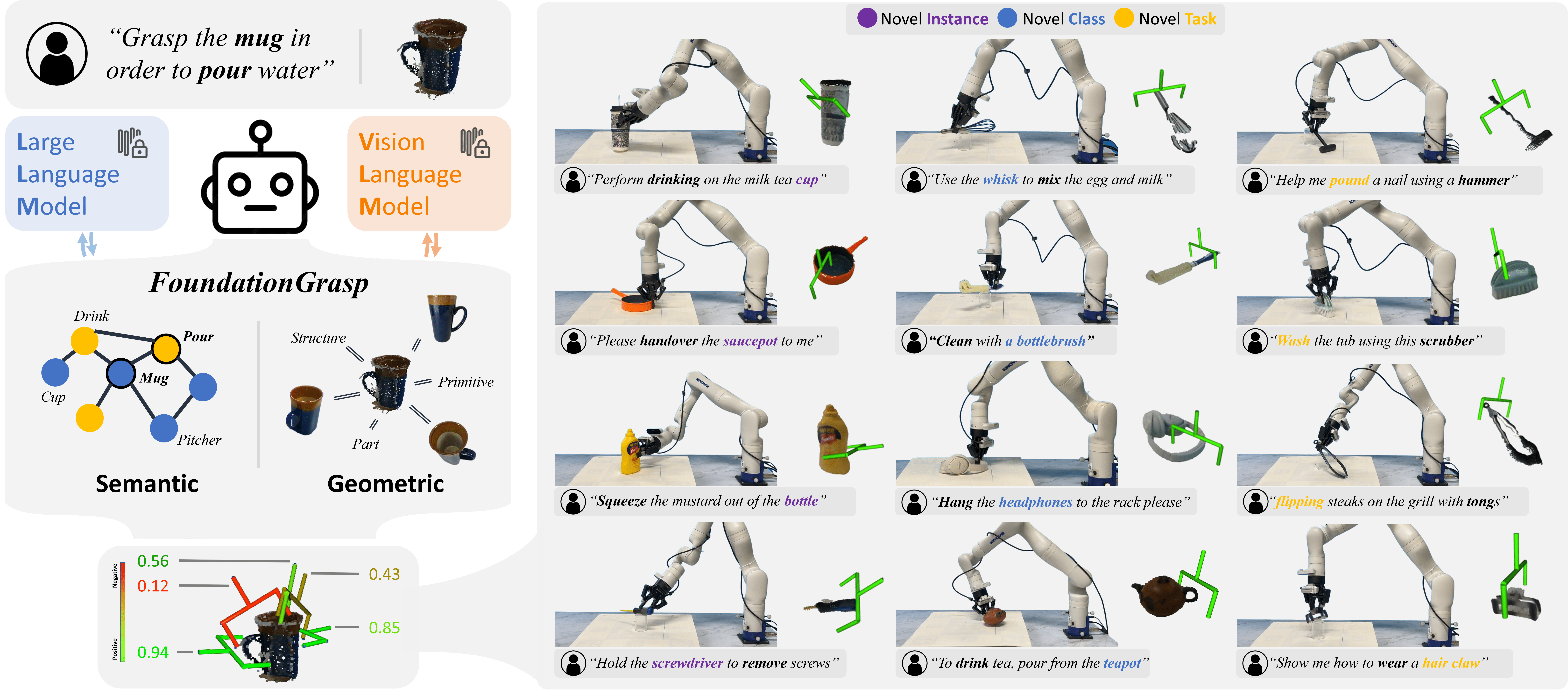}
  \captionof{figure}{FoundationGrasp leverages open-ended semantic and geometric knowledge from foundation models (i.e., LLMs and VLMs) to learn generalizable TOG skills. Perception and real-robot experiments demonstrate that FoundationGrasp can effectively generalize to novel object instances, object classes, and tasks unseen during training.}
  \label{fig:concept}
    \vspace*{-0.35in}
}

\makeatletter
\apptocmd{\@maketitle}{\insertfig}{}{}
\makeatother

\begin{document}

\title{FoundationGrasp: Generalizable Task-Oriented Grasping with Foundation Models}

\author{Chao Tang$^{1, 2}$, Dehao Huang$^{1, 2}$, Wenlong Dong$^{1, 2}$, Ruinian Xu$^{3}$, Hong Zhang$^{1, 2}$,~\IEEEmembership{Fellow,~IEEE}
        % <-this % stops a space
        
% \thanks{This paper was produced by the IEEE Publication Technology Group. They are in Piscataway, NJ.}% <-this % stops a space
% \thanks{Manuscript received April 19, 2021; revised August 16, 2021.}

% \thanks{This paper was produced by the IEEE Publication Technology Group. They are in Piscataway, NJ.}% <-this % stops a space
% \thanks{This paper was produced by the IEEE Publication Technology Group. They are in Piscataway, NJ.}% <-this % stops a space
% \thanks{Manuscript received April 19, 2021; revised August 16, 2021.}

\thanks{$^{1}$Shenzhen Key Laboratory of Robotics and Computer Vision, Southern University of Science and Technology, Shenzhen, China.}%
\thanks{$^{2}$Department of Electronic and Electrical Engineering, Southern University of Science and Technology, Shenzhen, China.}%
\thanks{$^{3}$Institute for Robotics and Intelligent Machines, Georgia Institute of
Technology, Atlanta, United States.}%

}

% The paper headers
\markboth{Journal of \LaTeX\ Class Files,~Vol.~14, No.~8, August~2021}%
{Shell \MakeLowercase{\textit{et al.}}: A Sample Article Using IEEEtran.cls for IEEE Journals}

% \IEEEpubid{0000--0000/00\$00.00~\copyright~2021 IEEE}
% Remember, if you use this you must call \IEEEpubidadjcol in the second
% column for its text to clear the IEEEpubid mark.

\maketitle

\begin{abstract}
  Task-oriented grasping (TOG), which refers to synthesizing grasps on an object that are configurationally compatible with the downstream manipulation task, is the first milestone towards tool manipulation. Analogous to the activation of two brain regions responsible for semantic and geometric reasoning during cognitive processes, modeling the intricate relationship between objects, tasks, and grasps necessitates rich semantic and geometric prior knowledge about these elements. Existing methods typically restrict the prior knowledge to a closed-set scope, limiting their generalization to novel objects and tasks out of the training set. To address such a limitation, we propose FoundationGrasp, a foundation model-based TOG framework that leverages the open-ended knowledge from foundation models to learn generalizable TOG skills. Extensive experiments are conducted on the contributed \underline{La}nguage and \underline{Vi}sion \underline{A}ugmented TaskGrasp (LaViA-TaskGrasp) dataset, demonstrating the superiority of FoundationGrasp over existing methods when generalizing to novel object instances, object classes, and tasks out of the training set. Furthermore, the effectiveness of FoundationGrasp is validated in real-robot grasping and manipulation experiments on a 7-DoF robotic arm. Our code, data, appendix, and video are publicly available at \href{https://sites.google.com/view/foundationgrasp}{https://sites.google.com/view/foundationgrasp}.

 % 

 % At the core of FoudationGrasp, we establish linguistic connections between novel and familiar objects and tasks with the help of semantic descriptions from a large language model (LLM). Meanwhile, to facilitate geometric understanding, a vision-language model (VLM) is employed to ground multi-view RGB images along with geometric descriptions to object point clouds. 

 % Our code, data, appendix, and video are publicly available at \textcolor{red}{xxxxxxx}.

\end{abstract}

 \def\abstractname{Note to Practitioners}

\begin{abstract}
    This research is motivated by the challenge of generalizable task-oriented grasping skill learning. Solving such a challenge could significantly improve the robot's level of automation and intelligence in tool manipulation for household and industrial tasks. Existing methods struggle with handling unseen objects and tasks in dynamic, open-world environments. To overcome this limitation, we propose to leverage the open-ended knowledge from foundation models to improve the generalization capabilities of existing TOG methods. This way, the robot can perform TOG w.r.t. unseen objects and tasks, facilitating downstream tool manipulation. Overall, this research has broad applicability to various scenarios involving tool manipulation, such as cleaning kitchenware and assembling parts in industrial contexts.
\end{abstract}

% The proposed method currently employs a sampling-and-evaluation pipeline, which relies heavily on the quality of the off-the-shelf grasp sampler. Future work will focus on developing a one-stage TOG pipeline to improve robustness and efficiency. 


\section{Introduction}
\IEEEPARstart{E}{ndowing} intelligent robots with tool manipulation capabilities, approaching human levels, could significantly expand their potential to perform a variety of household and industrial tasks, such as part assembly and kitchenware cleaning. As a fundamental step towards tool manipulation, task-oriented grasping (TOG) necessitates robots synthesizing grasps configurationally compatible with the downstream manipulation task. Task-incompatible grasping will inevitably limit the success of manipulation tasks or, in some cases, pose physical risks to the user (e.g., handover a knife by grasping the handle and exposing the sharp blade to the receiver). 

% In the context of kitchenware cleaning, for example, the household robot should grasp the handle of the brush to clean the stained coffee mug and then hang the cleaned mug on the rack by grasping its body or rim \cite{simeonov2022neural}. 

Despite the significance of TOG, such skills are not yet available to robots. Considering a household robot operating in open-world environments, such as apartments and garages, it needs to handle an open set of elements, including novel objects (instances, classes) and tasks unseen during training. A straightforward approach is to collect TOG data for each object with respect to each task in a brute-force manner. However, this approach is impractical due to the immense manpower and time required. In practical terms, one would expect training the robot only on a limited set of objects and tasks, while the robot is capable of generalizing the learned TOG skills to novel elements beyond the training examples.

% Second, in cognitive psychology, it is observed that two brain regions in the prefrontal cortex, which are respectively responsible for semantic and geometric reasoning, are jointly activated during cognitive processes\cite{wilson1993dissociation, huth2016natural}. Analogously, robots learning TOG skills requires semantic (e.g., ``\textit{what functions does a mug afford?}") and geometric (e.g., ``\textit{what geometric primitives a mug is composed of?}") understanding related to these open-set elements, which makes learning generalizable TOG skills even more challenging. 

% Consequently, it remains an open problem on generalizing TOG skills learned from a limited set of examples to open-set elements that are not necessarily trained with.
% Analogously, robots learning TOG skills requires semantic (e.g., ``\textit{what functions does a mug afford?}") and geometric (e.g., ``\textit{what geometric primitives a mug is composed of?}") understanding related to these open-set elements
% To achieve this goal, recent works have prposed to incorporate semantic and geometric knowledge into TOG pipelines. 

% add curretn situation
Prior work in TOG has approached this expectation through two main paths: the analytic approach and the data-driven approach. The former evaluates the task-oriented grasp quality with task wrench space analysis \cite{li1988task}. However, it assumes the availability of full object and hand models, which makes it less suitable for handling novel elements in open-world environments. Consequently, recent research has shifted towards the data-driven approach \cite{dang2012semantic, yang2019task, fang2020learning, qin2020keto}, which holds promise for better generalization capability. Especially, inspired by the observation that two brain regions in the prefrontal cortex, responsible for semantic and geometric reasoning, are jointly activated during cognitive processes\cite{wilson1993dissociation, huth2016natural}, more recent data-driven TOG methods \cite{montesano2008learning, song2010learning, murali2021same, detry2017task, antanas2019semantic} have integrated semantic and geometric knowledge into TOG pipelines to enhance adaptability. For example, Murali et al.\cite{murali2021same} contribute the largest and most diverse TOG dataset, TaskGrasp, and build a knowledge graph (KG) capturing semantic relationships between objects and tasks within the TaskGrasp dataset. Concurrently, several studies \cite{detry2017task, antanas2019semantic} consider both semantic and geometric properties of pre-defined sets of objects and tasks. Despite the successes of these methods, \textbf{they struggle to generalize to novel elements out of the pre-defined sets}. This limitation can be naturally attributed to the closed-set nature of the prior knowledge used. While such a limitation may be acceptable under the closed-world assumption, it results in sub-optimal TOG performance for household robots operating in open-world environments.
% , huth2016naturasl}

% In cognitive psychology, it is observed that two brain regions in the prefrontal cortex, which are respectively responsible for semantic and geometric reasoning, are jointly activated during cognitive processes\cite{wilson1993dissociation, huth2016natural}. Inspired by this observation, recent works \cite{montesano2008learning, song2010learning, murali2021same,abaci2005bridging, detry2017task, antanas2019semantic} have proposed incorporating semantic and geometric knowledge into TOG pipelines to improve the robots' adaptability to different situations. While the these methods have exhibited a degree of success, \textbf{they experience significant performance degradation when handling novel elements out of the pre-defined sets}. This can be naturally attributed to the fact that the supporting knowledge is limited to a closed-set scope. While such a limitation may be acceptable under the closed-world assumption, a household robot often operates in open-world environments, thus resulting in unsatisfying TOG performance in real-world applications.

% Montesano et al. \cite{montesano2008learning} and Song et al. \cite{song2010learning} construct semantic knowledge bases (KBs) with pre-defined sets of objects, tasks, and actions. Similarly, 

The preceding discussion raises a critical question: \textbf{How can we endow robots with open-ended knowledge about objects and tasks to learn generalizable TOG skills?} Recent advancements in foundation models, such as Large Language Models (LLMs) and Vision-Language Models (VLMs), have enabled robots to seamlessly extract and harness the open-ended knowledge embedded in these models for physically grounded tasks \cite{liang2023code, huang2023voxposer, huang2023visual, shah2023lm, ahn2022can}. Motivated by this trend, we propose FoundationGrasp, a foundation model-based TOG framework. As depicted in Figure \ref{fig:concept}, in contrast to previous works that restrict the prior knowledge to a closed-set scope, FoundationGrasp leverages the open-ended knowledge from foundation models to learn generalizable TOG skills.

% that are  to novel objects and tasks. 

% , which are trained on Internet-scale data, 

% \cite{hubel1965receptive} 

The pipeline of the proposed FoundationGrasp operates as follows. When presented with an object (e.g., \textit{mug}), a task (e.g., \textit{pour}), and a language instruction (e.g., \textit{``\textbf{pour} water from the \textbf{mug}"}), where the object and the task can be novel elements outside the training set, FoundationGrasp begins by prompting an LLM to generate semantic and geometric descriptions of the object and the task. Subsequently, an LLM-based semantic knowledge encoder and a VLM-based geometric knowledge encoder transform them with multi-modal sensory inputs into their feature representations. In the final stage, we draw inspiration from cognitive psychology and introduce a Transformer-based\cite{vaswani2017attention} task-oriented grasp evaluator with semantic and geometric branches. Evaluation on the contributed \underline{La}nguage and \underline{Vi}sion \underline{A}ugmented TaskGrasp (LaViA-TaskGrasp) dataset demonstrates that FoundationGrasp outperforms existing TOG methods when generalizing to novel object instances, object classes, and tasks beyond the training set. To validate the effectiveness in real-world applications, we further deploy FoundationGrasp on a Kinova Gen3 7-DoF robotic arm for task-oriented grasping and manipulation. 

% On the one hand, the semantic branch uses semantic descriptions to establish linguistic connections between the novel elements in the language instruction and familiar ones described during training. On the other hand, to facilitate the geometric understanding of objects and tasks, the geometric branch fuses multi-modal geometric features distilled from a VLM into the 3D representation.

This study builds on the preliminary work presented in GraspGPT \cite{tang2023graspgpt}, which showcases the potential of improving the generalization capability of TOG with an LLM in a proof-of-concept manner. GraspGPT primarily addresses the semantic aspect of TOG by leveraging language transferability to establish semantic connections between novel and familiar \textbf{linguistic} concepts (i.e., object class labels and task labels). However, semantic understanding alone is insufficient. Geometric understanding is also crucial for TOG since it enables the robot to \textbf{visually} identify different object parts and local geometries suitable for various tasks and affordances. For example, “spoon” and “saucepan” are semantically different but geometrically similar. Solely using semantic knowledge struggles to relate these concepts. However, both objects share the same geometric decomposition: a long handle plus a functional head. Building on this intuition, the extended version, FoundationGrasp, leverages both semantic and geometric knowledge from foundation models to learn generalizable TOG skills. Moreover, we include additional baselines, experiments, and studies to thoroughly demonstrate the value of incorporating foundation models in solving TOG problems and provide insights into the limitations. An extended discussion of related work on task-agnostic grasping, task-oriented grasping, and the applications of foundation models in robotics is also presented.

The main contributions of this paper are as follows: 
\begin{itemize}
    \item We propose FoundationGrasp, a foundation model-based TOG framework that leverages the open-ended knowledge from foundation models to learn generalizable TOG skills. 
    \item Built on top of the public TOG benchmark TaskGrasp dataset, we contribute a multi-modal TOG dataset, \underline{La}nguage and \underline{Vi}sion \underline{A}ugmented TaskGrasp (LaViA-TaskGrasp) dataset, to support the training and evaluation of generalizable TOG methods.
\end{itemize}
% \textcolor{red}{The rest of this paper is organized as follows.}

The rest of this paper is organized as follows. In Section \ref{related}, we review related work. Section \ref{problem_form} provides a problem formulation, and Section \ref{approach} describes the details of the proposed FoundationGrasp framework. In Section \ref{exp_setup} and Section \ref{exp}, the experimental setup, experiment results, and analysis are presented. Finally, we discuss the limitations in Section \ref{discuss} and conclude the paper in Section \ref{conclusion}.

    % \item A physical system is built to enable the robot to perform task-oriented grasping with novel objects and tasks guided by users' natural language instructions.

% Brown et al. [4], there are four key aspects to learning
% task-oriented tool usage: (a) understanding the desired effect,
% (b) identifying properties of an object that make it a suitable
% tool, (c) determining the correct orientation of the tool prior
% to usage, and (d) manipulating the tool

%  Tool use and learning
% in robots. In Encyclopedia of the Sciences of Learning,

\section{Related Work}\label{related}

\subsection{Task-Agnostic Grasping}
Grasping has been a fundamental problem in robotics. Earlier studies optimize grasp quality to output analytic solutions satisfying stability constraints, such as force closure and form closure \cite{ferrari1992planning}. However, a significant drawback of the analytic approach is its reliance on complete object and hand models \cite{tung1996fast}, which impedes its ability to generalize to novel objects.

% tung1996fast, prattichizzo2012manipulability, rosales2012synthesis

With the rise of machine learning techniques in recent years, the field of grasp synthesis and planning has been revolutionized by data-driven methods \cite{mahler2017dex, chu2018real, mousavian20196, lin2020using, sundermeyer2021contact} trained on various types of sensory data, including RGB-D images, depth images, and point clouds. For example, Dex-Net 2.0\cite{mahler2017dex} and DeepGrasp\cite{chu2018real} employ convolutional neural networks  (CNNs) to process RGB-D images and predict SE(2) planar grasp poses. PS-CNN \cite{lin2020using} takes advantage of both analytic and data-driven approaches by performing primitive shape decomposition on depth images to generate 6-DoF grasp poses in SE(3) space. \cite{mousavian20196, sundermeyer2021contact} approach the problem of grasp synthesis on point clouds with sampling-and-evaluation pipelines. While a substantial portion of research in robotic grasping focuses on task-agnostic grasping (TAG), the goal of grasping is often beyond this realm. Ignoring the task-oriented effect on object grasping limits the application of TAG methods in many cases. FoundationGrasp complements TAG methods by predicting grasps satisfying both stability and task-compatibility constraints.

% , wang2020neural

% Advanced robotic skills necessitate modeling complex relationships between objects, tasks, and grasps.

% PS-CNN \cite{lin2020using} takes advantage of both analytic and data-driven approaches by performing primitive shape decomposition on depth images to generate 6 DoF grasp poses in SE(3) space.

\subsection{Task-Oriented Grasping}
% classical and data-driven methods
In terms of TOG, where the grasping problem is contextualized in manipulation tasks, the objective is to grasp the object in a manner that is configurationally compatible with the downstream manipulation task. Classical TOG methods evaluate the task-oriented grasp quality with task wrench space analysis \cite{li1988task}. However, they share similar limitations with earlier studies in TAG due to the inherent diversity of objects and tasks in open-world environments. Recently, data-driven methods have achieved success in solving TOG problems to some extent. Dang et al.\cite{dang2012semantic} build a task-oriented grasp planner that considers partial object geometry, tactile contacts, and hand kinematic data. Yang et al.\cite{yang2019task} expand previous works to enable TOG in object-stacking scenes. To alleviate the burden of data collection, Fang et al.\cite{fang2020learning} and Qin et al. \cite{qin2020keto} train task-dependent TOG models with simulated self-supervision. Nevertheless, the aforementioned methods derive object-task-grasp relationships solely from training data without external knowledge sources and consequently exhibit limited generalization performance. Diffusion models have emerged as potential tools for TOG \cite{barad2023graspldm, hao2024language, ju2024robo}. However, their generalization capabilities remain questionable. 

% \cite{haschke2005task, li1988task, prats2007task}.

% \cite{kokic2020learning} learns TOG skills from human activities.

% Semantic -> KB-based -> affordance -> Semantic + Geometric -> FM-related
% Since learning TOG skills necessitates semantic and geometric understanding of objects and tasks, 
More recent works have incorporated external prior knowledge into TOG pipelines. Montesano et al.\cite{montesano2008learning} and Song et al. \cite{song2010learning}\cite{song2015task} construct semantic knowledge bases (KBs) with pre-defined sets of objects, tasks, actions, and constraints. Similarly, Ardón et al. \cite{ardon2019learning} use Markov Logic Networks (MLN) to build a semantic KB, encoding objects, tasks, visual attributes, and indoor locations of 30 commonly found household objects. To enrich a KB with a broader range of objects and tasks, Murali et al.\cite{murali2021same} contribute the largest and most diverse TOG dataset, TaskGrasp, and a resulting KG. More importantly, they propose the state-of-the-art graph-based TOG method, GCNGrasp, to generalize to objects and tasks within the KG. In addition to only incorporating semantic knowledge, a series of works \cite{detry2017task, antanas2019semantic} consider both semantic and geometric properties of pre-defined sets of objects and tasks. For example, Detry et al.\cite{detry2017task} approach the TOG problem with a geometric grasp model and a CNN-based semantic model. Similarly, in \cite{antanas2019semantic}, a visual module that estimates object geometric properties is integrated with a probabilistic logic module. Another line of research utilizes affordance knowledge to predict task-oriented grasps\cite{xu2021affordance, nguyen2023language, liu2020cage}. Xu et al. \cite{xu2021affordance} combine affordance segmentation and keypoint detection to guide task-oriented grasp prediction. Liu et al.\cite{liu2020cage} propose a novel TOG representation with part affordances, part material, object states, and tasks.

% \cite{gibson1977theory}

% Do et al.\cite{do2018affordancenet} detect 9 pre-defined affordance classes as priors to build correspondences between tasks and grasps. 

While these methods have demonstrated the generalization capability to some extent, they suffer from generalization to novel objects and tasks beyond pre-defined sets. This phenomenon can be explained by the fact that these methods limit the prior knowledge to a closed-set scope. To learn generalizable TOG skills, we propose to leverage open-ended knowledge from foundation models in this work.

% , and use Bayesian networks to capture the statistical dependencies between them. 

\subsection{Foundation Models in Robotics}
% Foundation models, such as LLMs\cite{brown2020language, thoppilan2022lamda, chowdhery2022palm} and VLMs\cite{radford2021learning, li2022blip},
% , spanning from masked language modeling\cite{DBLP:journals/corr/abs-1810-04805} to zero-shot image classification\cite{li2021align}.
Foundation models, such as LLMs and VLMs, encode open-ended knowledge about the world and have exhibited impressive capabilities on a wide range of visual and linguistic tasks. In robotics, such knowledge is beneficial for physically grounded tasks, such as tabletop manipulation\cite{liang2023code, huang2023voxposer}, navigation\cite{huang2023visual, shah2023lm}, and mobile manipulation\cite{ahn2022can}. While most of them primarily rely on foundation models for task planning and scene understanding, FoundationGrasp grounds the open-ended knowledge to the grasp synthesis process. 

Regarding robotic grasping, LERF-TOGO\cite{sharma2023language} builds a Language Embedded Radiance Fields (LERF) with features distilled from a VLM to produce a 3D relevancy heatmap given an object part name. LAN-Grasp \cite{mirjalili2023lan} first prompts an LLM to output an object part name appropriate for grasping and uses a VLM to identify the corresponding part in the image. These methods\cite{sharma2023language, mirjalili2023lan, li2024shapegrasp} focus on solving ``where to grasp" and ignore ``how to grasp". FoundationGrasp optimizes 6-DoF task-oriented grasp poses to jointly address ``where to grasp" and ``how to grasp". GraspCLIP\cite{tang2023task} and ATLA\cite{ren2023leveraging} extract knowledge from foundation models for manipulation and grasping. However, they are limited to 2D planar grasping. In contrast, FoundationGrasp studies TOG in 3D space.

% 3DAPNet \cite{nguyen2023language} utilizes task word embeddings to build connections between novel and familiar tasks, avoiding being limited to a pre-defined set of tasks. We will later experimentally prove that FoundationGrasp better extracts and harnesses the open-end knowledge embedded in foundation models compared to naively relying on task word embeddings.

% Nguyen et al.\cite{nguyen2023language} propose 3DAPNet, a 6 DoF task-oriented grasp planner which utilizes task word embeddings to build connections between different tasks, avoiding being limited to a predefined set of tasks. 

% open-end LERF-TOGO, Lan-Grasp (only where no how)
% Leveraging (semantic and geometric knowledge about object classes from LLM, only 4 DoF)
% GraspCLIP 
% 3DAPNet

\begin{table*}[t]
\centering
\renewcommand\arraystretch{1.3}
\setlength\tabcolsep{15pt}%调列距
  % \vspace*{-0.1in}
\begin{tabular}{ccc}
\toprule
\textbf{Class} & \textbf{O2O Description}                                                                                                                                             & \textbf{O2T Description}                                                                                                                                                                                             \\ \hline \specialrule{0em}{3pt}{3pt}
\textit{Mug}   & \begin{tabular}[c]{@{}c@{}}{[}sim\_geo{]} \textit{``Objects such as teacups, jars, glasses, and} \\ \textit{cylindrical vases have similar geometries to a mug."}\end{tabular}   & \begin{tabular}[c]{@{}c@{}}{[}func{]} \textit{``A mug is a typically cylindrical household object with a} \\ \textit{handle, used primarily for drinking hot beverages, such} \\ \textit{as coffee, tea, or hot chocolate."}\end{tabular} \\ \specialrule{0em}{3pt}{3pt} \hline
\textbf{Task}  & \textbf{T2T Description}                                                                                                                                             & \textbf{T2O  Description}                                                                                                                                                                                            \\ \hline \specialrule{0em}{3pt}{3pt}
\textit{Sweep} & \begin{tabular}[c]{@{}c@{}}{[}sim\_effect{]} \textit{``Verbs such as clear, clean, brush, wipe, or} \\ \textit{dust achieve similar effects to 'sweep an object'."}\end{tabular} & \begin{tabular}[c]{@{}c@{}}{[}afford{]} \textit{``Brooms, dustpans, vacuum cleaners, and sweepers are} \\ \textit{household objects that afford the function of sweeping."}\end{tabular}                                        \\ \specialrule{0em}{3pt}{3pt} \bottomrule
\end{tabular}
  % \vspace*{0.1in}
\caption{Examples of semantic descriptions in the LaViA-TaskGrasp dataset}
\label{tab:example_sem_desc}
  % \vspace*{-0.1in}
\end{table*}

\begin{table*}[t]
\centering
\renewcommand\arraystretch{1.3}
\setlength\tabcolsep{13pt}%调列距
\begin{tabular}{cccc}
\toprule
\textbf{Class} & \textbf{O2P Description}                                                                                                                                                           & \textbf{Task}  & \textbf{T2P Description}                                                                                                                                                                                                           \\ \hline  \specialrule{0em}{3pt}{3pt}
\textit{Mug}   & \begin{tabular}[c]{@{}c@{}}{[}part{]} \textit{``A mug is composed of a cylindrical body, a} \\ \textit{handle, a circular base, and often features an} \\ \textit{outer glaze or design."}\end{tabular} & \textit{Sweep} & \begin{tabular}[c]{@{}c@{}}{[}primitive{]} \textit{``Primitive shapes on a household object that} \\ \textit{can be used for sweeping include the long, cylindrical} \\ \textit{handle and the fan-shaped bundle of bristles on a broom."}\end{tabular} \\ \specialrule{0em}{3pt}{3pt} \bottomrule
\end{tabular}
  % \vspace*{0.1in}
\caption{Examples of geometric descriptions in the LaViA-TaskGrasp dataset}
\label{tab:example_geo_desc}
  \vspace*{-0.2in}
\end{table*}

\section{Problem Formulation}\label{problem_form}
We consider the problem of learning a task-oriented grasping policy $\pi$ that takes in an input tuple $\gamma = (X_o, I_o, L)$ and outputs a task-oriented grasp pose $g^*$:
\begin{equation}
    g^* = \pi(\gamma) = \pi(X_o, I_o, L)
\end{equation}
where $X_o \in \mathbb{R}^{N \times 3}$ and $I_o \in \mathbb{R}^{H \times W \times 3}$ are the (partial) point cloud and the RGB image of the object instance $o$, respectively. $N$ is the number of points. $H$ and $W$ are the height and width of the image. $L$ is a natural language instruction specifying an object class $c$ and a task $t$, such as ``\textit{\textbf{pour} water with the \textbf{mug}}". $g^*$ is a 6-DoF grasp pose of a parallel-jaw gripper represented by $(R, T)$, where $R \in SO(3)$ represents the 3D orientation and $T \in  \mathbb{R}^{3}$ represents the 3D translation. 

We assume access to a TOG dataset $\mathcal{D}$, consisting of $K_o$ object instances, $\mathcal{O} = \{o_k\}_{k=1}^{K_o}$, with ground truth task-oriented grasp pose annotations. $\mathcal{D}$ spans a set of object classes $\mathcal{C} = \{c_i\}_{i=1}^{K_c}$ and a set of tasks $\mathcal{T} = \{t_j\}_{j=1}^{K_t}$, where $K_c$ and $K_t$ are the numbers of object classes and tasks in $\mathcal{D}$, respectively. We use $\mathcal{D}$ to supervise the training of $\pi$. During inference, $o$, $c$ and $t$ can be novel elements out of $\mathcal{D}$. The key challenge here is to generalize $\pi$ learned on $\mathcal{D}$ to an open set of elements not necessarily trained with. 

Mathematically, the goal is to estimate the posterior distribution $P(G|X_o, I_o, L)$, where $G$ represents the space of all task-oriented grasp poses. We factorize the estimation process into two steps: (1) task-agnostic grasp sampling $P(G|X_o, I_o)$ and (2) task-oriented grasp evaluation $P(S | X_o, I_o, L, g)$, where $S$ is the task-compatibility score of each candidate $g \in G$. Finally, the robot executes $g$ with the highest score (i.e., $g^*$).

% (a) An overview of GraspGPT framework: when presented with a novel concept, such as a novel object class or task, in the natural language
%  instruction, GraspGPT first prompts an LLM to acquire a set of language description paragraphs of the concept. Subsequently, GrasGPT evaluates the
%  task compatibility of grasp candidates based on the multi-modal inputs from the sensors and an LLM. (b) The detailed structure of task-oriented grasp
%  evaluator: the module is a customized transformer decoder that injects semantic knowledge from an LLM into the natural language instruction

\begin{figure*}[t]
  \centering
  \vspace*{-0.2in}
  \begin{tikzpicture}[inner sep = 0pt, outer sep = 0pt]
    \node[anchor=south west] (fnC) at (0in,0in)
      {\includegraphics[height=3.5in,clip=true,trim=0in 0in 0in 0in]{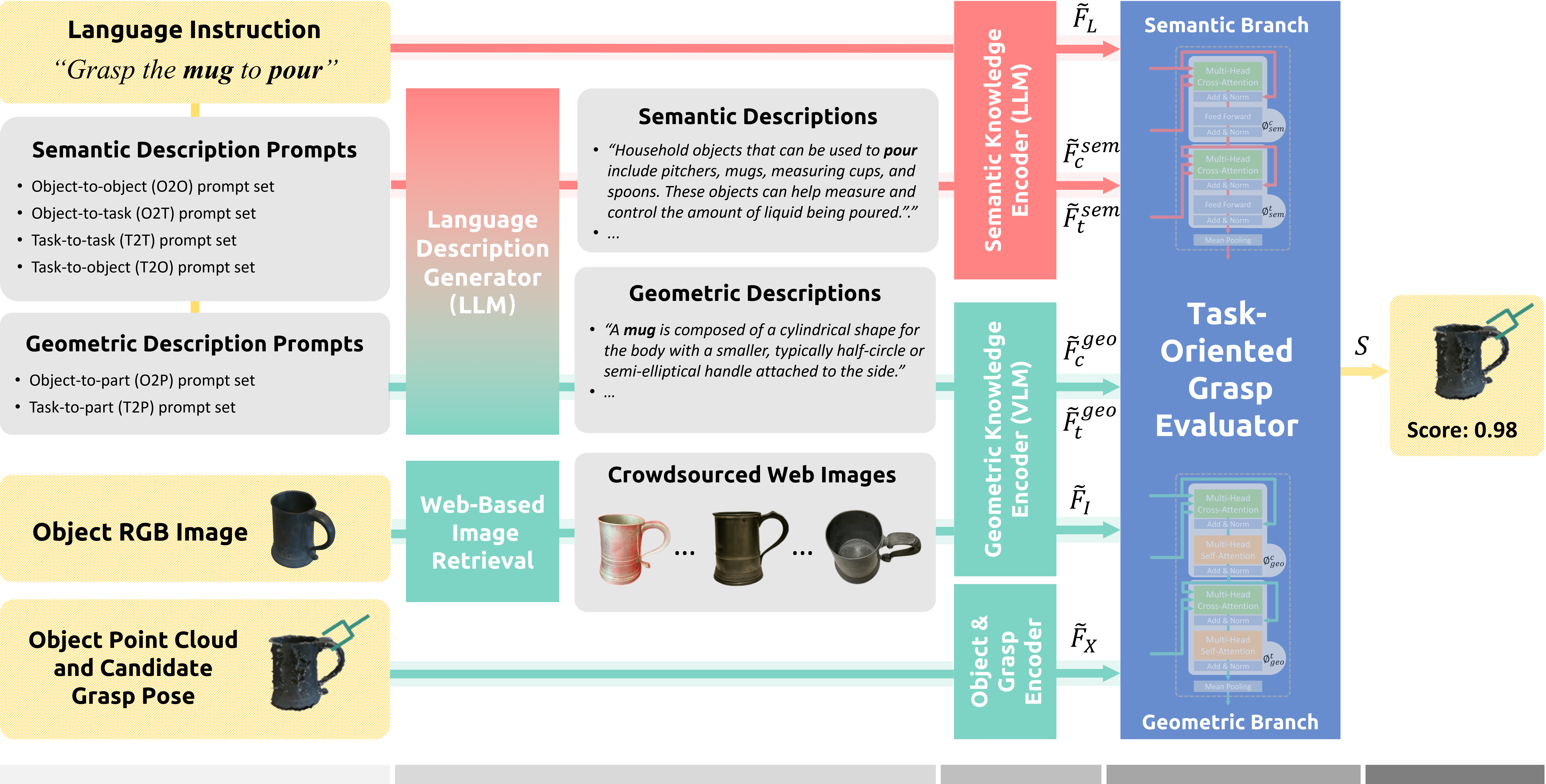}};
  \end{tikzpicture}
    % \vspace*{-0.1in}
  \caption{\textbf{An overview of FoundationGrasp framework:} The pipeline consists of (1) knowledge generation, (2) multi-modal feature representation, and (3) task-oriented grasp evaluation. When presented with a language instruction, FoundationGrasp first prompts an LLM to generate semantic and geometric descriptions of the object and the task. A web-based image retrieval module crowdsources images from the Internet. Subsequently, an LLM-based semantic knowledge encoder and a VLM-based geometric knowledge encoder transform them with multi-modal sensory inputs into their latent space feature representations. In the final stage, a Transformer-based task-oriented grasp evaluator with semantic and geometric branches evaluates the task compatibility of each grasp candidate.}
  \vspace*{-0.2in}
  \label{fig:pipeline}
\end{figure*}

% an object, a task, and 

% When presented with an object (e.g., mug), a task (e.g.,
% pour), and a language instruction (e.g., “pour water from the
% mug”), where the object and the task can be novel elements
% out of the training set, FoundationGrasp first prompts an
% LLM to generate semantic and geometric descriptions of the
% object and the task. Subsequently, an LLM-based semantic
% knowledge encoder and a VLM-based geometric knolwdge
% encoder transform them along with multi-modal sensory inputs
% into their feature representations. In the final stage, we draw
% inspirations from cognitive psychology [18] and introduce
% a Transformer-based [19] task-oriented grasp evaluator with
% semantic and geometric branches.

\section{FoundationGrasp}\label{approach}

% add algorithm

% \textcolor{red}{talk about cognitive science}

% escribe the geometric appearance of 3D point
% clouds, and it is relatively easy to obtain a simple text descrip-
% tion for a given 3D shape. It has been demonstrated in [54],
% [55], [58], [59] that it is feasible to use a text description to
% improve the understanding of point clouds.

\subsection{Overview}

% the lack of geometric cues and the partial sparseness

% This dedicated
% dataset focuses on capturing descriptive information concerning
% geometric parts within 3D shapes

% 
% why not use BLIP or GPT-4V to decribe the geometric properties of objects

% It has been demonstrated in [54],
% [55], [58], [59] that it is feasible to use a text description to
% improve the understanding of point clouds

% why geometric descriptions are not fine-grained, global structure information -> object part,  geometric assumption category-level 

An overview of the proposed  FoundationGrasp framework is depicted in Figure \ref{fig:pipeline}. The pipeline consists of three stages: (1) knowledge generation, (2) multi-modal feature representation, and (3) task-oriented grasp evaluation. In Section \ref{data_generation}, FoundationGrasp generates linguistic data, visual data, and grasp data based on $o$, $c$, and $t$ with foundation models or pre-trained models. The second stage described in Section \ref{feat_rep} transforms multi-modal data into their latent space feature representations. Finally, in Section \ref{task_grasp_eval}, a Transformer-based task-oriented grasp evaluator computes the task-compatibility score $S$ of each candidate grasp pose $g$. 

% Each stage will be introduced in detail for the rest of this section.

\subsection{Knowledge Generation}\label{data_generation}
The knowledge generation stage constructs the LaViA-TaskGrasp dataset by augmenting the TaskGrasp dataset with linguistic and visual data. The original TaskGrasp dataset consists of $K_o$ object instances, spanning $K_c$ object classes and $K_t$ tasks. The knowledge generation pipeline consists of four phases: (1) language description generation, (2) language instruction generation, (3) web-based image retrieval, and (4) grasp candidate generation.

% During inference, we follow the similar procedure to generate open-end knowledge when presented with novel elements out of LaViA-TaskGrasp dataset. 

% reformulate description generation

\textit{Language Description Generation} \  To extract open-ended knowledge from foundation models, we start by prompting an LLM to generate descriptions for each object class $c_i$ and task $t_j$ in the TaskGrasp dataset. We divide language description generation into semantic and geometric aspects and carefully design the LLM prompting strategy to capture the relationships between objects, tasks, and grasps. 

Semantic descriptions linguistically connect novel objects and tasks with familiar ones. Previous methods \cite{murali2021same, montesano2008learning, song2010learning, song2015task, ardon2019learning} primarily model three types of semantic relationships: (1) object-object, (2) task-task, and (3) object-task. We follow a similar spirit to design semantic description prompts. For each object class $c_i$, we design two prompt sets: (1) object-to-object (O2O) prompt set, with each prompt querying object classes sharing a similar property with $c_i$ (e.g., ``\textit{Describe what household objects have similar $P_{obj}$ to $c_i$}:"). Here, $P_{obj}$ represents object properties, such as shape, function, and meta-category; (2) object-to-task (O2T) prompt set, with each prompt querying tasks afforded by $c_i$ (e.g., ``\textit{Describe the common use of household object $c_i$}:"). Similarly, for each task $t_j$, we design (1) task-to-task (T2T) prompt set, with each prompt querying tasks that are semantically or physically relevant to $t_j$ (e.g., ``\textit{Describe what verbs are semantically close to $t_j$}:"); (2) task-to-object (T2O) prompt set, with each prompt querying object classes that afford $t_j$ (e.g., ``\textit{Describe what household objects can be used to $t_j$}:"). 

Geometric descriptions focus on two aspects: (1) the geometric structure of $c_i$ and (2) the correlations between geometric structures (form) and $t_j$ (function). To address the former, we design an object-to-part (O2P) prompt set, with each prompt querying the geometric structure of $c_i$ (e.g., ``\textit{Describe what parts the household object $c_i$ is composed of:}"). In parallel, for each task $t_j$, we design a task-to-part (T2P) prompt set, with each prompt querying geometric structures that support $t_j$ (e.g., ``\textit{Describe what geometric primitives on a household object can be used to $t_j$}:"). We regularize the outputs of the LLM to bullets of parts/primitive shapes. This regularization strategy reduces the difficulty of grounding geometric descriptions to 3D representations during task-oriented grasp evaluation and meanwhile prevents the LLM from hallucinating.

Each prompt set includes $N_p$ prompts, with $N_p$ varying for each set. To account for the inherent randomness of generated descriptions, we query the LLM $N_a$ times per prompt. Examples of generated semantic and geometric descriptions are showcased in Table \ref{tab:example_sem_desc} and \ref{tab:example_geo_desc}. Since a single description only offers a partial ``observation" of $c_i$ or $t_j$, we randomly combine descriptions from each prompt to construct complete semantic description paragraphs, denoted as $L_{c_i}^{sem}$ and $L_{t_j}^{sem}$, and geometric description paragraphs, denoted as $L_{c_i}^{geo}$ and $L_{t_j}^{geo}$. Typically, each paragraph has an approximate length of 2-6 sentences. A complete list of prompt sets and more examples of language descriptions can be found in the appendix. During inference, when presented with novel objects and tasks out of the LaViA-TaskGrasp dataset, FoundationGrasp follows the same procedure to generate semantic and geometric descriptions for them. Such an open-ended nature is the primary advantage of FoundationGrasp over previous TOG methods.

\textit{Language Instruction Generation} \ During training, the language goal is constructed from instruction templates derived from  \cite{nguyen2022affordance}. Each template in the initial set, such as ``\textit{Use the [obj] to [task]}", is populated with an object class $c_i$ and a task $t_j$. Recognizing that human-expressed language goals exhibit significant diversity, we further enrich the vocabulary and grammatical diversity of the initial template set using the LLM. Specifically, we prompt the LLM to perform template augmentation (e.g., ``\textit{rewrite the following sentence in a different grammatical format:}"), which leads to an augmented set of instruction templates. The augmentation process generates some unconventional formulations (e.g., ``\textit{ensure you have a [task]-friendly grip on the [obj]}"). While not commonly used in daily communication, we keep them to enhance the robustness to novel instructions. A complete list of instruction templates is available in the appendix.

% \begin{figure}[t]
%   \centering
%   % \vspace*{-0.2in}
%   \begin{tikzpicture}[inner sep = 0pt, outer sep = 0pt]
%     \node[anchor=south west] (fnC) at (0in,0in)
%       {\includegraphics[height=1.8in,clip=true,trim=0in 0in 0in 0in]{imgs/real-robot/multi-view-crowdsourced.png}};
%   \end{tikzpicture}
%     % \vspace*{-0.28in}
%   \caption{\textcolor{red}{multi-view images crowdsourcing}}
%   \label{fig:xxx}
%   % \vspace*{-0.25in}
% \end{figure} 

\textit{Web-Based Image Retrieval} \ Previous works in 3D scene representation have demonstrated that visual-language features distilled from 2D foundation models facilitate geometric understanding of 3D world \cite{kobayashi2022decomposing, thomason2022language}. Following the generation of language descriptions, we augment the TaskGrasp dataset by utilizing the Internet as a large-scale database to retrieve visual data. Each instance in the original TaskGrasp dataset is captured from only 2-3 viewpoints. Therefore, for each object instance $o_k$ in the dataset, we first query Google Cloud Vision API with its original images in the dataset, retrieving candidate images with respective confidence scores. Each candidate image contains an object instance sharing similar visual characteristics and geometric structures. Subsequently, we employ the Segment Anything Model (SAM) \cite{kirillov2023segment} for background removal. To guarantee the view diversity, we utilize CLIP \cite{radford2021learning} visual encoder to extract the embedding vector of each candidate image and compute the cosine similarities between them. Those with high similarities are excluded. Finally, a set of $N_I$ images (original + retrieved images) for $o_k$, denoted as $I_{o_k}$, is selected based on confidence scores. Examples of retrieved images are available in the appendix. Note that the goal here is not to perform accurate 3D reconstruction of $o_k$. Thus, the instances in the retrieved images do not need to match $o_k$ precisely. This image retrieval strategy aims to provide visual information for constructing geometric priors of objects and tasks in the later stage. During inference, the robot utilizes $I_o$ captured by the onboard camera.

% Previous studies in point cloud shape completion \cite{aiello2022cross, song2023fine} have demonstrated that structure concepts learned
% from multi-view 2D images facilitates the geometric understanding of 3D point clouds.

\textit{Grasp Candidate Generation} \ Each object instance $o_k$ in the TaskGrasp dataset is annotated with 25 candidate grasp poses and task labels. These annotations supervise the training of $\pi$. During inference, we first apply Contact-GraspNet \cite{sundermeyer2021contact} to generate $N_g^+$ collision-free grasp candidates on $X_o$. $N_g^+$ varies depending on the specific scene. Then, a subset of $N_g$ grasps ($N_g \leq N_g^+$) is selected using farthest point sampling (FPS) to construct the final set of candidate grasps. The selection of $N_g$ is a trade-off between grasp diversity and density.  

% Task-agnostic grasp candidate sampling is formulated as $P(G|X_o, I_o)$.

% 30 -100, 50

\begin{figure}[t]
  \centering
  \vspace*{-0.2in}
  \begin{tikzpicture}[inner sep = 0pt, outer sep = 0pt]
    \node[anchor=south west] (fnC) at (0in,0in)
      {\includegraphics[height=2.2in,clip=true,trim=0.2in 0in 0in 0in]{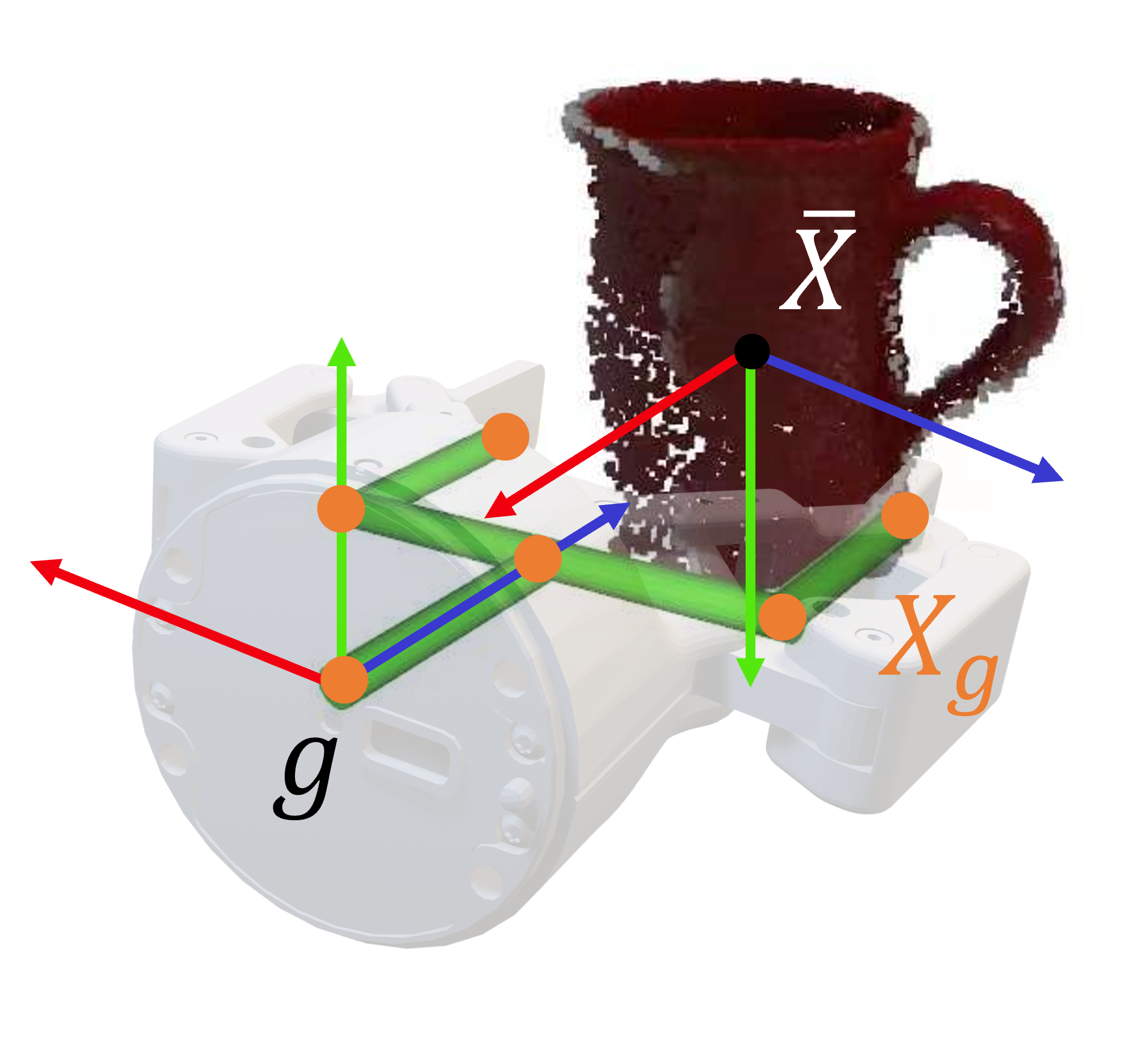}};
  \end{tikzpicture}
    \vspace*{-0.2in}
  \caption{A grasp $g$ is represented with six control points $X_g$ on the gripper model in the object reference frame. The origin of the object frame is $\overline{X}$, the center of mass of the object point cloud $X_o$.}
  \label{fig:grasp_rep}
  \vspace*{-0.2in}
\end{figure} 

% Grasps are defined
% in the object reference frame, whose origin is X¯, the center
% of mass of the observed point cloud.

\subsection{Multi-Modal Feature Representation}\label{feat_rep}
To incorporate open-ended knowledge from the previous stage into the FoundationGrasp framework, we transform them with multi-modal sensory inputs into their latent space feature representations. Three encoders are introduced: an object and grasp encoder, an LLM-based semantic knowledge encoder, and a VLM-based geometric knowledge encoder. 

% Each of which will be introduced in detail.

% using an extra binary feature that indicates whether a point
% belongs to the object or to the gripper

% point cloud + grasp -> PN++
\textit{Object and Grasp Encoder} \ To model the relative spatial relationship between the gripper and the object, the first step involves transforming $g \in SE(3)$ and $X_o \in \mathbb{R}^{N \times 3}$ into the same space. Following the variational grasp generator 6-DoF GraspNet \cite{mousavian20196}, the first step is achieved by approximating $g$ with six control points, represented as $X_g \in \mathbb{R}^{6 \times 3}$ in the object frame (shown in Figure \ref{fig:grasp_rep}). Then, $X_o$ and $X_g$ are jointly embedded. Specifically, $X_o$ and $X_g$ are first concatenated to form $X_{o+g} \in \mathbb{R}^{(N+6) \times 3}$. A binary feature vector is appended to indicate whether a point belongs to $X_o$ or $X_g$. Subsequently, a point encoder (denoted as \textnormal{PN++}) extracts localized features from $X_{o+g}$, which associates features to a reduced number of $N^-$ points ($N^- < N$):
\begin{align}
    X_{o+g} &= \textnormal{Concat}([X_o, X_g], \textnormal{dim}=0) \\ F_X & =  \textnormal{PN++}(X_{o+g})
\end{align}
where point embeddings $F_X \in \mathbb{R}^{N^- \times 1024}$. PN++ is adopted from PointNet++\cite{qi2017pointnet++}, a deep learning architecture designed for directly processing 3D point cloud data without losing the inherent structure and details. We choose to maintain a degree of locality in the latent space, as opposed to relying on a single global embedding, to preserve the localized geometric information. Such information is necessary for understanding the relationship between geometric structures, objects, and tasks, as discussed in Section \ref{data_generation}. The point encoder consists of three set abstraction (SA) layers with multi-scale grouping (MSG). A Multi-Layer Perceptron (MLP) is then applied for dimension reduction to output $\Tilde{F}_X \in \mathbb{R}^{N^- \times 256}$.

% and \textnormal{PN++} denotes the point encoder

 % understanding the relative spatial relationship between the gripper and the object at specific spatial point.

% $F_X$ is later fed to the geometric branch for task-oriented grasp evaluation.  

% capture localized features of the input point cloud

% ummarized at a small number of
% points/pixels

% keep a degree of locality, i.e., associating features to a small number of points N X ă N rather than a
% single global embedding because the information about the missing part that needs to be estimated
% by the entire model is also mostly localized.

% have a sufficiently
% large receptive field to infer some global information about the object.

% reduce the cardinality of the point cloud. Pooling has the double purpose of
% expanding the receptive field to also include more global information and reduce the complexity of
% the subsequent cross-attention operations fusing the two modalities.

% \begin{center}
% ``\textit{Objects such as spatulas, ladles, and spoons are kitchen utensils that support the function of scoop.}"
% \end{center}

% A common choice is training a dedicated language encoder from scratch based on language descriptions generated in Section \ref{data_generation}. However, it would take substantial training effort and, meanwhile, achieve limited generalization to novel linguistic concepts unseen during training.

% , such as CLIP language encoder, 

\textit{Semantic Knowledge Encoder}  \ To linguistically connect novel and familiar objects and tasks, FoundationGrasp needs to interpret a broad spectrum of linguistic concepts present in semantic descriptions. Consider, for instance, a T2O description of task ``\textit{scoop}": ``\textit{Objects such as spatulas, ladles, and spoons are kitchen utensils that support the function of scoop.}" In this context, FoundationGrasp needs to identify object classes (e.g., \textit{spatula}, \textit{ladle}, \textit{spoon}), understand their categorical taxonomy (e.g., \textit{utensil}), and recognize associated tasks (e.g., \textit{scoop}). Training a dedicated language encoder from scratch would take substantial training effort and, meanwhile, achieve limited generalization to novel linguistic concepts unseen during training. As an alternative, we utilize a large pre-trained BERT\cite{devlin2018bert} to encode $L_{c_i}^{sem}$ and $L_{t_j}^{sem}$. Other large pre-trained language models would also suffice. These LLMs are pre-trained on a large corpus of text data so that they can interpret and generalize to novel linguistic concepts without any fine-tuning. Specifically, the pre-trained BERT processes a paragraph from $L_{c_i}^{sem}$ to output $F_{c_i}^{sem} \in \mathbb{R}^{T_{c} \times 768}$, and a paragraph from $L_{t_j}^{sem}$ to output $F_{t_j}^{sem} \in \mathbb{R}^{T_{t} \times 768}$. Likewise, we transform a language instruction $L$ into $F_L \in \mathbb{R}^{T_L \times 768}$. $T_{c}$, $T_{t}$, and $T_L$ denote the maximum lengths (with zero-padding) of $L_{c}^{sem}$, $L_{t}^{sem}$, and $L$, respectively. We further project language sequence embeddings to a lower dimension and obtain $\Tilde{F}_{c_i}^{sem} \in \mathbb{R}^{T_{c} \times 512}$, $\Tilde{F}_{t_j}^{sem} \in \mathbb{R}^{T_{t} \times 512}$, and $\Tilde{F}_L \in \mathbb{R}^{T_L \times 512}$.

% We process semantic description paragraphs (i.e., $L_{c_i}^{sem}$ or $L_{t_j}^{sem}$) by transforming them to $F_{c_i}^{sem} \in \mathbb{R}^{T_{os} \times 768}$ and $F_{t_j}^{sem} \in \mathbb{R}^{T_{ts} \times 768}$

% \textcolor{red}{remove BERT equations!}

% The processing of transforming semantic description paragraphs (i.e., $L_{c_i}^{sem}$ or $L_{t_j}^{sem}$) can be represented as below:
% \begin{equation*}
%         F_{c_i}^{sem}, F_{t_j}^{sem} = \textnormal{BERT}(L_{c_i}^{sem}, L_{t_j}^{sem})
% \end{equation*}
% where $F_{c_i}^{sem} \in \mathbb{R}^{T_{os} \times 768}$ and $F_{t_j}^{sem} \in \mathbb{R}^{T_{ts} \times 768}$. 

\textit{Geometric Knowledge Encoder}  \ Geometric knowledge is twofold, including images and geometric descriptions. Images visually capture the geometric structure of an object, while geometric descriptions provide a linguistic interpretation of the object's geometry. Similar to previous works \cite{kobayashi2022decomposing, thomason2022language}, we extract multi-modal geometric features with CLIP, which is pre-trained on a large number of geometric concepts (e.g., primitives, parts, and objects) in the form of image-text pairs. During each training loop, we first extract intermediate CLIP visual features of a random image from $I_{o_k}$ to obtain $F_{I_{o_k}} \in \mathbb{R}^{h \times w \times 1024}$. Paragraphs from $L_{c_i}^{geo}$ and $L_{t_j}^{geo}$ are then processed by the CLIP language encoder to obtain $F_{c_i}^{geo} \in \mathbb{R}^{1024}$ and $F_{t_j}^{geo} \in \mathbb{R}^{1024}$, respectively. Then, flattening and dimension reduction are performed to output $\Tilde{F}_{I_{o_k}} \in \mathbb{R}^{hw \times 128}$, $\Tilde{F}_{c_i}^{geo} \in \mathbb{R}^{128}$, and $\Tilde{F}_{t_j}^{geo} \in \mathbb{R}^{128}$.

\begin{figure*}[t]
  \centering
  \vspace*{-0.3in}
  \begin{tikzpicture}[inner sep = 0pt, outer sep = 0pt]
    \node[anchor=south west] (fnC) at (0in,0in)
      {\includegraphics[height=3.2in,clip=true,trim=0in 0.5in 0in 0.5in]{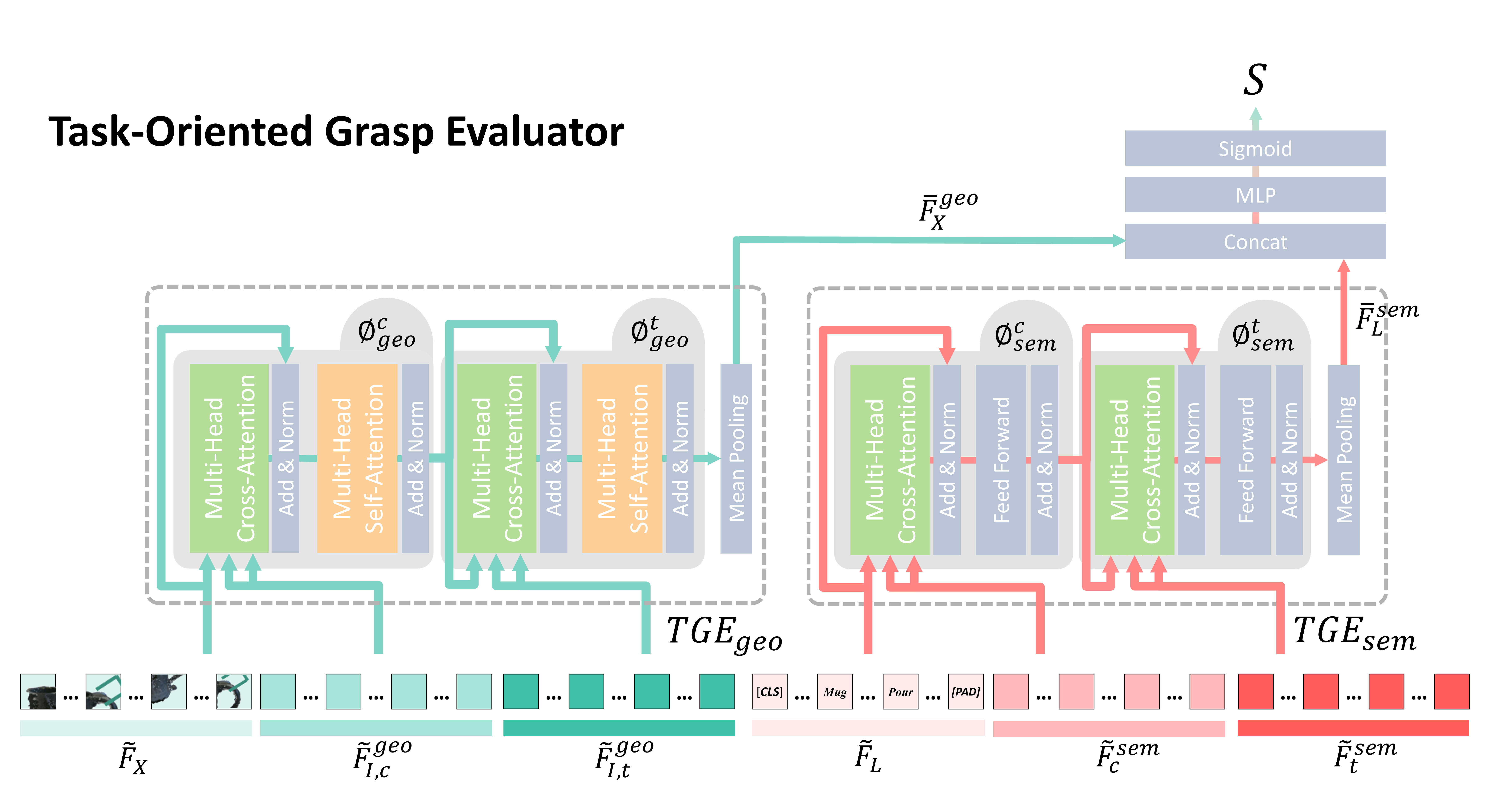}};
  \end{tikzpicture}
    % \vspace*{0.05in}
  \caption{Task-oriented grasp evaluator is a customized Transformer consisting of a geometric branch  $TGE_{geo}$ (left) and a semantic branch $TGE_{sem}$ (right).}
  \label{fig:sem_geo_branch}
  \vspace*{-0.25in}
\end{figure*} 

\subsection{Task-Oriented Grasp Evaluation}\label{task_grasp_eval}
The final stage evaluates the task compatibility of $g$, leveraging the semantic and geometric knowledge embedded in multi-modal features. To this end, we propose a Transformer-based task-oriented grasp evaluator with semantic and geometric branches. For simplicity, we drop the subscripts of multi-modal features in the following discussions. Formally, task-oriented grasp evaluation can be formulated as:
\begin{align}
    S = \textnormal{TGE}(\Tilde{F}_X, \Tilde{F}_{I}, \Tilde{F}_{c}^{geo}, \Tilde{F}_{t}^{geo}, \Tilde{F}_L, \Tilde{F}_{c}^{sem}, \Tilde{F}_{t}^{sem})
\end{align}
where TGE represents the task-oriented grasp evaluator, and $S$ is the task-compatibility score of $g$. The first four inputs are processed by the geometric branch $\textnormal{TGE}_\textnormal{geo}$, and the latter three are processed by the semantic branch $\textnormal{TGE}_\textnormal{sem}$. 

\textit{Semantic Branch} \ $\textnormal{TGE}_\textnormal{sem}$ aims to linguistically connect novel objects and tasks with familiar ones using semantic descriptions. The problem is analogous to machine translation in natural language processing (NLP), which aims to transform language sequences from the source to the target domain. In our context, the robot initially cannot understand novel elements in $L$. $\textnormal{TGE}_\textnormal{sem}$ translates them to familiar ones using semantic descriptions. Inspired by this similarity, we design $\textnormal{TGE}_\textnormal{sem}$ based on the Transformer decoder, an effective architecture for handling sequence-to-sequence problems.

The translation process is divided into two phases: object knowledge incorporation and task knowledge incorporation, with each phase being addressed by a Transformer decoder layer. The architecture of $\textnormal{TGE}_\textnormal{sem}$ is presented in Figure \ref{fig:sem_geo_branch} (right), which consists of $\phi^{sem}_{c}$ and $\phi^{sem}_t$. Each layer incorporates semantic description embeddings $\Tilde{F}^{sem}$ (i.e., $\Tilde{F}_{c}^{sem}$ or $\Tilde{F}_{t}^{sem}$) into $\Tilde{F}_L$:
\begin{align}
    \phi^{sem}(\Tilde{F}^{sem}, \Tilde{F}_L) \rightarrow \Tilde{F}_L^{sem} \in \mathbb{R}^{T_L \times 512}
\end{align}
where $\Tilde{F}_L^{sem}$ denotes language instruction embeddings contextualized with semantic knowledge. The computational procedure of the two layers is similar, which can be represented as:
  \vspace*{-0.2in}
\begin{subequations}\label{eq:decoder_layer}
\begin{align}
    \Tilde{F}_L^{sem} &= \text{LN}(\Tilde{F}_L + \text{MHA}(\Tilde{F}_L, \Tilde{F}^{sem})) \\
    \Tilde{F}_L^{sem} &= \text{LN}(\Tilde{F}_L^{sem}+ \text{FFN}(\Tilde{F}_L^{sem})
\end{align}
\end{subequations}
where LN, MHA, and FFN denote layer normalization, multi-head attention, and feedforward networks, respectively. MHA consists of $h_{sem}$ heads. Each head reconstructs $\Tilde{F}_L$  with elements in $\Tilde{F}^{sem}$ weighted by their normalized correspondence: 
  \vspace*{-0.2in}
\begin{subequations}
\begin{align}
    \text{head}_i &= \text{Attention}(Q_L, K_{sem}, V_{sem}) \\
    &= \text{Softmax}(\frac{Q_L K_{sem}^T}{\sqrt{d_k}})V_{sem}
\end{align}
\end{subequations}
where $n = 1, 2, ..., h_{sem}$. $Q_L$, $K_{sem}$, and $V_{sem}$ can be computed as:
\begin{align}
    Q_L = \Tilde{F}_LW_n^Q, K_{sem} = \Tilde{F}^{sem}W_n^K, V_{sem} = \Tilde{F}^{sem}W_n^V
\end{align}
where $W_n^Q \in \mathbb{R}^{512 \times d_k}$, $W_n^K \in \mathbb{R}^{512 \times d_k}$, $W_n^V \in \mathbb{R}^{512 \times d_v}$, and $d_k = d_v$. During training, semantic descriptions are associated with corresponding object and task tokens, while irrelevant information is disregarded. Finally, $\Tilde{F}_L^{sem}$ is mean pooled to obtain the language instruction embedding $\overline{F}_L^{sem} \in \mathbb{R}^{128}$  contextualized with semantic knowledge.

% &= \text{Attention}(Q_L, K_{sem}, V_{sem}) \\

% The computation within each head is as follows:

% , each of which can be represented as:
% \begin{align}
%     \text{MHA}(\Tilde{F}_L, \Tilde{F}^{sem}) &= \text{Concat}[head_1, ..., head_{sem}]W^O 
% \end{align}
% where $W^O \in \mathbb{R}^{h_{sem}d_v \times 512}$.

% Formally, each layer models the following distribution:
% \begin{align}
%         P(\Tilde{F}_L^{sem}|\Tilde{F}^{sem}, \Tilde{F}_L) = P(z_1, ..., z_{T_L} | x_1, ..., x_{T_{*}}; y_1, ..., y_{T_L})
% \end{align}
% % = P(Z | x_1, ..., x_{T_{*}}; y_1, ..., y_{T_L}) \\
% where $T_{*}$ can be $T_c$ or $T_t$. 

% where $\Tilde{F}_L^{sem}$ denotes language instruction embeddings contextualized with semantic knowledge.

% $\mathbb{R}^{T_{L} \times 512} \times \mathbb{R}^{T_{*d} \times 512} \rightarrow \mathbb{R}^{T_{L} \times 512}$,

% Intuitively, the weights determine the extent to which each step’s context “focuses” on each input token, and the key is to make this process for assigning the weights differentiable so that it can be learned along with all of the other neural network parameters.

%  In translation tasks, attention models often assigned high attention weights to cross-lingual synonyms when generating the corresponding words in the target language.

% Both training
% and inference follow the same computational procedure

% linguistic connections

% \begin{align}
%     \phi^{sem}(\Tilde{F}^{sem}, \Tilde{F}_L) \rightarrow \Tilde{F}_L^{sem} \in \mathbb{R}^{T_L \times 512}
% \end{align}

\textit{Geometric Branch} \ To facilitate the geometric understanding of objects and tasks, $\textnormal{TGE}_\textnormal{geo}$ fuses multi-modal geometric features distilled from 2D VLM to 3D. Specifically, to match the spatial dimension of $\Tilde{F}_{I}$, the geometric description embeddings are first repeated at each spatial location to obtain $\Tilde{F}_{c}^{geo} \in \mathbb{R}^{hw \times 128}$ and $\Tilde{F}_{t}^{geo} \in \mathbb{R}^{hw \times 128}$. Then, multi-modal geometric features are constructed as below:
\begin{align}
    \Tilde{F}_{I}^{geo} = \text{Concat}([\Tilde{F}_{I}, \Tilde{F}^{geo}], \text{dim=-1})
\end{align}
where $\Tilde{F}_{I}^{geo} \in \mathbb{R}^{hw \times 256}$ ($\Tilde{F}_{I, c}^{geo}$ or $\Tilde{F}_{I, t}^{geo}$ ). The architecture of $\textnormal{TGE}_\textnormal{geo}$ is depicted in Figure \ref{fig:sem_geo_branch} (left). Similar to the semantic branch, the geometric branch is divided into object and task knowledge incorporation, which correspond to $\phi_{c}^{geo}$ and $\phi_{t}^{geo}$ in $\textnormal{TGE}_\textnormal{geo}$, respectively. Each layer incorporates geometric features $\Tilde{F}^{geo}_{I}$ into $\Tilde{F}_X$:
\begin{align}
    \phi^{geo}(\Tilde{F}^{geo}_{I}, \Tilde{F}_X) \rightarrow \Tilde{F}_X^{geo} \in \mathbb{R}^{N^- \times 256}
\end{align}
The two layers share a similar design, where geometric features are fused with a cross-attention layer and rectified by a self-attention layer:
\begin{subequations}
\begin{align}
    \Tilde{F}_X^{geo} &= \textnormal{LN}(\Tilde{F}_X + \textnormal{MHA}(\Tilde{F}_X, \Tilde{F}_{I}^{geo} )) \\
    \Tilde{F}_X^{geo} &= \textnormal{LN}(\Tilde{F}_X^{geo} + \textnormal{MHA}(\Tilde{F}_X^{geo}))
\end{align}
\end{subequations}
where MHA consists of $h_{geo}$ heads. The computational procedure of cross-attention fusion is similar to Equation \ref{eq:decoder_layer}. The self-attention rectification operates similarly to cross-attention fusion, with the distinction that query, key, and value tensors are all projections of $\Tilde{F}_X^{geo}$. The design choice, where a rectification layer follows a fusion layer, is based on the consideration that a self-attention layer performs a permutation-invariant transformation to rectify geometric knowledge features that are not properly aligned\cite{zhang2021view}. Finally, $\Tilde{F}_X^{geo}$ is processed by an SA layer with global pooling to output the point embedding $\overline{F}_X^{geo} \in \mathbb{R}^{300}$ contextualized with geometric knowledge.

% $\textnormal{TGE}_\textnormal{geo}$ concludes with an additional cross-attention layer that integrates features across different levels of abstraction. 

% The design of $\textnormal{TGE}_\textnormal{geo}$ draws inspiration from the domain of point cloud shape completion, which necessitates strong prior knowledge about geometric structures. Previous studies in shape completion have demonstrated that image \cite{zhang2021view, aiello2022cross} and text \cite{song2023fine} are two important modalities for providing such prior knowledge. To facilitate the geometric understanding of objects and tasks, we propose to fuse multi-view RGB images and geometric descriptions to $F_X$.

% Subsequently, we are ready to establish correspondences between point features and multi-modal geometric knowledge features. 

% Then, we concatenate $\Tilde{F}_{o}$ and $\Tilde{F}^{geo}$ (i.e., $\Tilde{F}_{c}^{geo}$ or $\Tilde{F}_{t}^{geo}$) channel-wise, which can be represented as:
% \begin{align*}
%     F_{vl}^{geo} &= \textnormal{Concat}([\Tilde{F}_{o}, \Tilde{F}^{geo}], \textnormal{dim}=-1)
% \end{align*}
% where $\Tilde{F}_{vl}^{geo} \in \mathbb{R}^{hw \times 256}$. 

Finally, the language instruction embedding $\overline{F}_L^{sem}$ is concatenated with the point embedding $\overline{F}_X^{geo}$ to compute $S$, which can be represented as:
\begin{align}
    S = \sigma(\textnormal{MLP}(\textnormal{Concat}([\overline{F}_X^{geo}, \overline{F}_L^{sem}]), \textnormal{dim=-1})))
\end{align}
where $\sigma$ denotes the sigmoid activation function.

\textit{Loss Function} \ The binary cross-entropy loss between $S$ and the ground truth label $S_{gt}$ is computed as below:
\begin{subequations}
\begin{align}
    \mathcal{L} &= \textnormal{BCE}(S, S_{gt}) \\
    &= -\frac{1}{M} \sum_{m=1}^{M} S_{gt, m} \cdot \text{log}(S_{m}) + (1 - S_{gt, m}) \cdot \text{log}(1-S_{m})
\end{align}
\end{subequations}
where $M$ is the number of samples, and $S_{gt, m}$ is set to one if the $m_{th}$ grasp pose is compatible with $t$ and zero otherwise.

\section{Experimental Setup}\label{exp_setup}
In this section, we evaluate the performance of FoundationGrasp against existing baselines under three different held-out settings, including held-out instance setting, held-out (object) class setting, and held-out task setting. We divide the experiments into two parts: perception experiments and real-robot experiments. While the former evaluates the performance on the LaViA-TaskGrasp dataset, the latter is conducted on a real robot. Lastly, implementation details for model design and training are provided.

\subsection{Perception Experiments}

\textit{Baselines}  \ We compare FoundationGrasp to the following baselines that are representative of existing methods:
\begin{itemize}
    \item \textbf{TAG} represents task-agnostic grasping methods \cite{mahler2017dex, chu2018real, mousavian20196, lin2020using, sundermeyer2021contact} that focus on satisfying the stability constraint while ignoring the task-compatibility constraint. Specifically, we use Contact-GraspNet\cite{sundermeyer2021contact} as our baseline.
    
    \item \textbf{SGN} is an adapted version of Semantic Grasp Network \cite{liu2020cage}. It learns object-task-grasp relationships solely from training data without incorporating external knowledge.

    \item \textbf{SGN+we} is an enhanced version of SGN. It uses ConceptNet\cite{speer2017conceptnet} pre-trained word embeddings as prior knowledge to initialize object classes and tasks. 
    
    \item \textbf{GCNGrasp} \cite{murali2021same} is the state-of-the-art graph-based TOG method. It utilizes a semantic KG connecting a pre-defined set of objects and tasks. In our implementation, if an object class or task is not present in the graph, we connect the node to its nearest neighbor in the graph by measuring the cosine similarity of their ConceptNet pre-trained word embeddings. Specifically, we compare FoundationGrasp with two versions of GCNGrasp: the open-world version (GCNGrasp-ow) and the closed-world version (GCNGrasp-cw). The open-world version adheres to the same train/test split as other baselines, while the closed-world version assumes access to all (i.e., training and test) object classes and tasks in its KG, along with their ground truth relations. Although this assumption is impractical for real-world robotic applications, it allows us to highlight the performance gap between generalizing to elements within versus outside the pre-defined sets. By default, GCNGrasp refers to GCNGrasp-ow unless specified otherwise.
    
    % \item \textbf{GraspGPT} \cite{tang2023graspgpt} is the preliminary version of FoundationGrasp. Essentially, it focuses on the semantic understanding of objects and tasks using semantic descriptions generated by an LLM. 
\end{itemize}
We also report the performance of \textbf{GraspGPT} to highlight the technical extension from the preliminary version.

\begin{figure}[t]
  \centering
  % \vspace*{-0.3in}
  \begin{tikzpicture}[inner sep = 0pt, outer sep = 0pt]
    \node[anchor=south west] (fnC) at (0in,0in)
      {\includegraphics[height=1.55in,clip=true,trim=0in 0in 0in 1in]{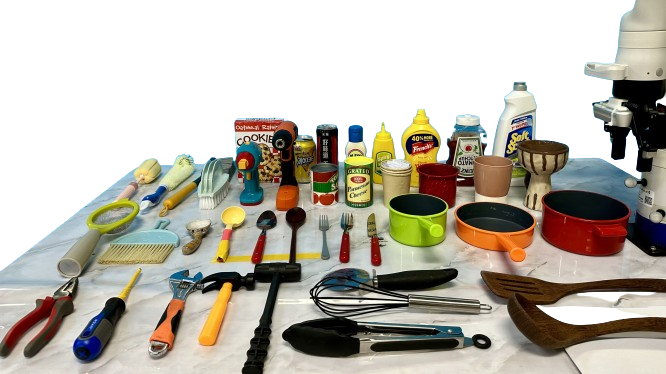}};
  \end{tikzpicture}
    % \vspace*{-0.1in}
  \caption{Novel objects tested in real-robot experiments cover commonly used accessories, kitchen utensils, and mechanic tools.}
  \label{fig:test_objects}
  \vspace*{-0.2in}
\end{figure} 

\textit{Dataset}  \ FoundationGrasp and baselines are evaluated on the contributed LaViA-TaskGrasp dataset, which includes 56 tasks, 75 object classes, and 191 object instances. Each object instance is a point cloud of a real household object reconstructed from \textbf{multiple viewpoints}. 25 candidate grasp poses with binary labels (1 for task-compatible, 0 for task-incompatible) are annotated on each object instance. For each held-out setting, $\mathcal{K}$-fold cross validation ($\mathcal{K}=4$) is performed. In each fold, 25\% of instances/classes/tasks are used for testing, and the remaining 75\% are used for training and validation. In terms of language data, the LaViA-TaskGrasp dataset contains 40 geometric descriptions, 80 semantic descriptions for each object class, and 40 geometric descriptions, 50 semantic descriptions for each task. We randomly combine these descriptions to generate 750 object semantic/geometric description paragraphs and 560 task semantic/geometric description paragraphs. LaViA-TaskGrasp dataset also includes 53 language instruction templates, resulting in a diverse set of over two million possible language instructions. In terms of vision data, we prepare $N_I=24$ images, as in \cite{zhang2021view}, for $o_k$ following the procedure described in Section \ref{data_generation}.

\textit{Metrics}  \ According to the convention in previous works\cite{murali2021same, tang2023graspgpt}, we compute the Average Precision (AP) scores for each object instance, object class, and task. These scores are then averaged over all instances, classes, and tasks to obtain instance mAP (mean AP), class mAP, and task mAP.

\subsection{Real-Robot Experiments}
\textit{Single-Object Task-Oriented Grasping Experiments} \ The experiment platform consists of a 7-DoF Kinova Gen3 robotic arm equipped with a Robotiq 2F-85 parallel jaw gripper and an Intel RealSense D435 RGB-D camera with eye-in-hand calibration. As shown in Figure \ref{fig:test_objects}, we collect 42 test objects unseen during training from the YCB dataset and our laboratory, with variations in functionality, use case, geometry, and material. The test objects cover commonly used accessories, kitchen utensils, and mechanic tools. During each trial, a test object is randomly selected and placed on a flat-surface kitchen table with varying poses. The robot then receives a language instruction from the user specifying the target object and the desired task. During inference, SAM is first applied to extract the test object from the scene point cloud captured from a single view. Contact-GraspNet \cite{sundermeyer2021contact} then samples  $N_g^+$  grasp candidates for evaluation. Finally, the robot picks the grasp candidate with the highest score for execution. Compared to the \textbf{multi-view} setting in perception experiments, the \textbf{single-view} setup is used in real-robot experiments since a robot, in most cases, only has a partial observation of the object.

% Shown in Figure \ref{fig:test_objects}, we collect test objects from the YCB dataset and our laboratory, covering commonly used accessories, kitchen utensils, and mechanic tools. During inference, SAM \cite{kirillov2023segment} is first applied to extract the test object from the point cloud captured from a single view. 

\textit{Multi-Object Task-Oriented Grasping Experiments}  \ The multi-object TOG experiments are conducted at three levels of clutter: (1) light clutter, where objects are spaced apart and clearly visible without occlusion in the camera view; (2) medium clutter, where light occlusions and contacts exist between objects with some functional parts not fully visible; (3) heavy clutter, where significant physical contacts and heavy occlusions occur between objects or functional parts. For each trial, a random set of 3 to 10 objects is selected, including distractor objects out of the test object set (e.g., apple, plate). Detection and segmentation are performed using GroundingDINO \cite{liu2023grounding} and SAM. The rest of the setup remains consistent with that used in the single-object TOG experiments.

\textit{Task-Oriented Manipulation Experiments} \ To support the hypothesis that FoundationGrasp facilitates downstream tool manipulation, we combine FoundationGrasp with pre-programmed manipulation skills and conduct task-oriented manipulation experiments. Presented in Table \ref{tab:mani_quantitative}, we conduct manipulation experiments on 9 household tasks selected from the household chores list \cite{yamazaki2012home}. Since manipulation policy learning is not the focus of this paper, we design rule-based heuristics to determine the operating direction of the tool object and compute the effect point on the receptacle object. Manipulation trajectories are generated by merging sequential motion primitives.

\textit{Metrics} \ The single-object TOG pipeline is divided into three stages: Perception, Planning, and Action. We report the statistics for each stage separately, providing details for failure analysis. 100 trials are conducted on the held-out instance, class, and task settings, with 10 per instance, class, or task. A trial is considered successful if the robot correctly detects TOG poses, plans the end effector to the top-ranked TOG pose, and physically executes the pose on the target object. Since there are no ground truth TOG labels for test objects, task compatibility is determined manually by the user. Stability is verified if the object is grasped and lifted for three seconds by the robot. In the multi-object TOG experiments, we additionally report the success of object grounding, referred to as “Grounding,” to indicate whether the robot correctly selects the target object. If multiple objects meet the language instruction criteria (e.g., two mugs for drinking), successfully executing TOG poses on any of them is considered a success. 15 trials are conducted at each clutter level, resulting in 45 trials in total. For task-oriented manipulation experiments, we record the success rates for both task-oriented grasping and manipulation. Each of the 9 household tasks is tested with 10 trials.

\subsection{Implementation Details}
Training and inference are conducted on a desktop PC with a single NVIDIA RTX 3090 GPU. All the baselines are re-trained and tested on the LaViA-TaskGrasp dataset with the same loss function for fair comparison. FoundationGrasp is implemented using Pytorch 1.11.0 with CUDA 11.3. We use an Adam \cite{kingma2014adam} optimizer with a weight decay of 0.0001. The learning rate is set to 0.0001 initially and adjusted with a LambdaLR scheduler as in \cite{murali2021same, tang2023graspgpt}.  To increase the robustness to unseen object poses, input point clouds are augmented with random scaling, rotation, jitter, and dropout. Each point cloud is then downsampled to 4096 points and fed into FoundationGrasp. The model is trained for 50 epochs with a batch size of 32.

For the choice of foundation models, we use OpenAI GPT-4 to generate language descriptions and OpenAI CLIP implemented by OpenCLIP as the VLM. For the large pre-trained language encoder, we choose the Google pre-trained BERT-Base model provided by Hugging Face. All the foundation models are frozen during training. To facilitate the training efficiency, the knowledge described in Section \ref{data_generation} is pre-generated and pre-processed with selected foundation models before training. More details on model training and design can be found in the appendix. 

For real-robot experiments, after determining the target grasp pose, we use MoveIt!\cite{chitta2012moveit} for trajectory generation. Specifically, KDL\cite{beeson2015trac} and RRT-Connect\cite{kuffner2000rrt} are used for inverse kinematics computation and trajectory planning, respectively. Collision checking is performed using the FCL\cite{pan2012fcl} package. Upon reaching the target grasp pose, pre-defined gripper aperture settings and maximum torque are executed by the controller of the Robotiq 2F-85 gripper.

\begin{figure*}[t]
  \centering
  % \vspace*{-0.2in}
  \begin{tikzpicture}[inner sep = 0pt, outer sep = 0pt]
    \node[anchor=south west] (fnC) at (0in,0in)
      {\includegraphics[height=2.6in,clip=true,trim=0in 0in 0in 0in]{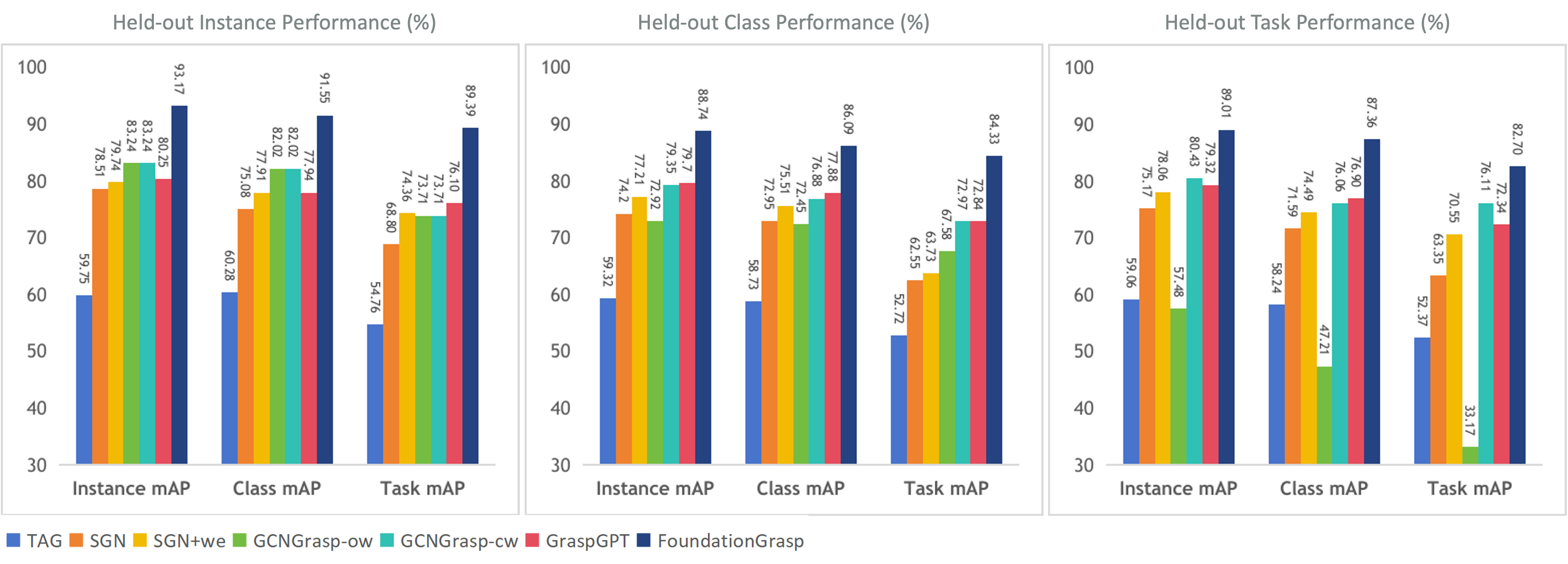}};
  \end{tikzpicture}
    % \vspace*{-0.1in}
  \caption{Quantitative results of perception experiments}
  % \vspace*{-0.1in}
  \label{fig:perception_quantative}
\end{figure*}

\begin{figure*}[t]
  \centering
  % \vspace*{-0.2in}
  \begin{tikzpicture}[inner sep = 0pt, outer sep = 0pt]
    \node[anchor=south west] (fnC) at (0in,0in)
      {\includegraphics[height=3.2in,clip=true,trim=0.5in 0in 0in 0in]{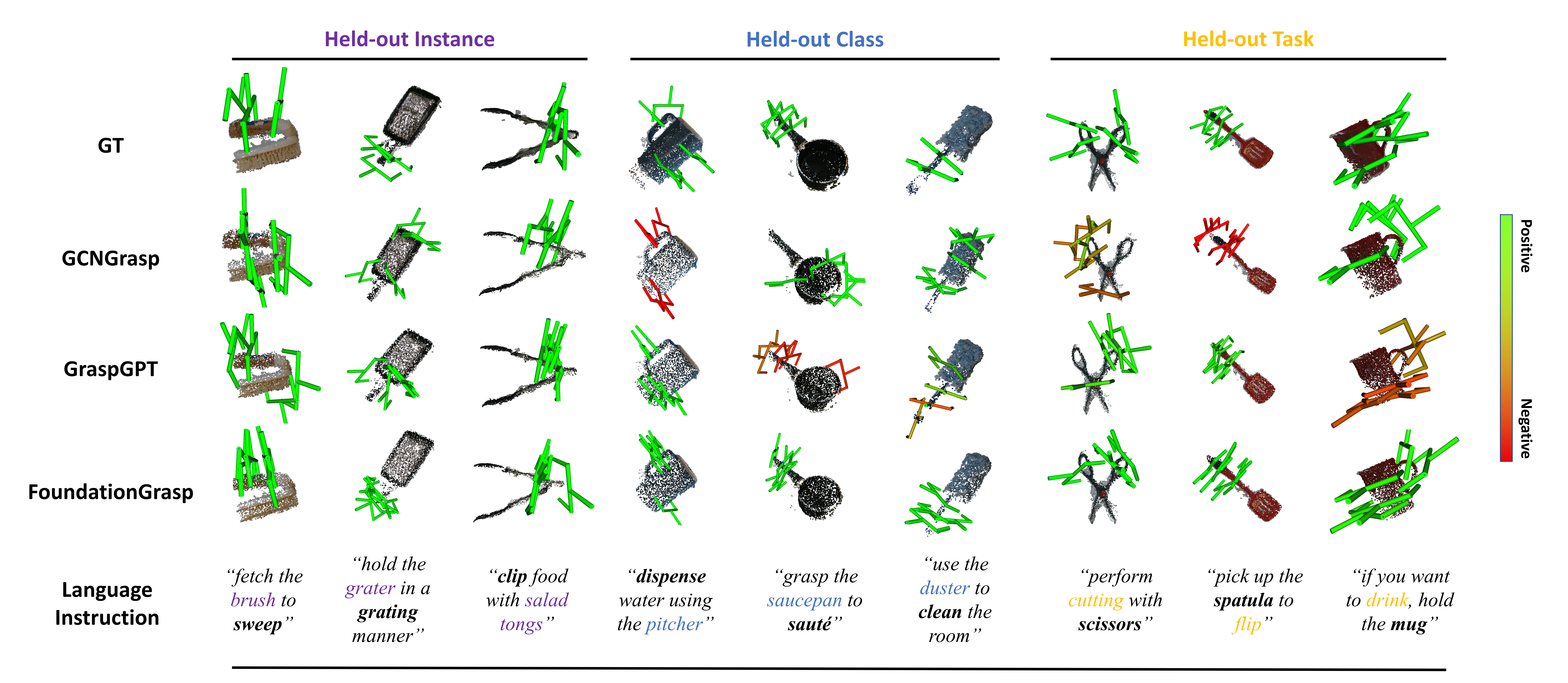}};
  \end{tikzpicture}
    % \vspace*{-0.28in}
  \caption{Qualitative results of perception experiments. GT denotes ground truth task-oriented grasp poses. Top-5 predictions of each method are visualized. Grasp poses are colored by their confidence scores (green is higher).}
  \vspace*{-0.1in}
  \label{fig:grasp_qualitative}
\end{figure*} 

\begin{figure}[t]
  \centering
  % \vspace*{-0.1in}
  \begin{tikzpicture}[inner sep = 0pt, outer sep = 0pt]
    \node[anchor=south west] (fnC) at (0in,0in)
      {\includegraphics[height=1.7in,clip=true,trim=0in 0in 0in 0.3in]{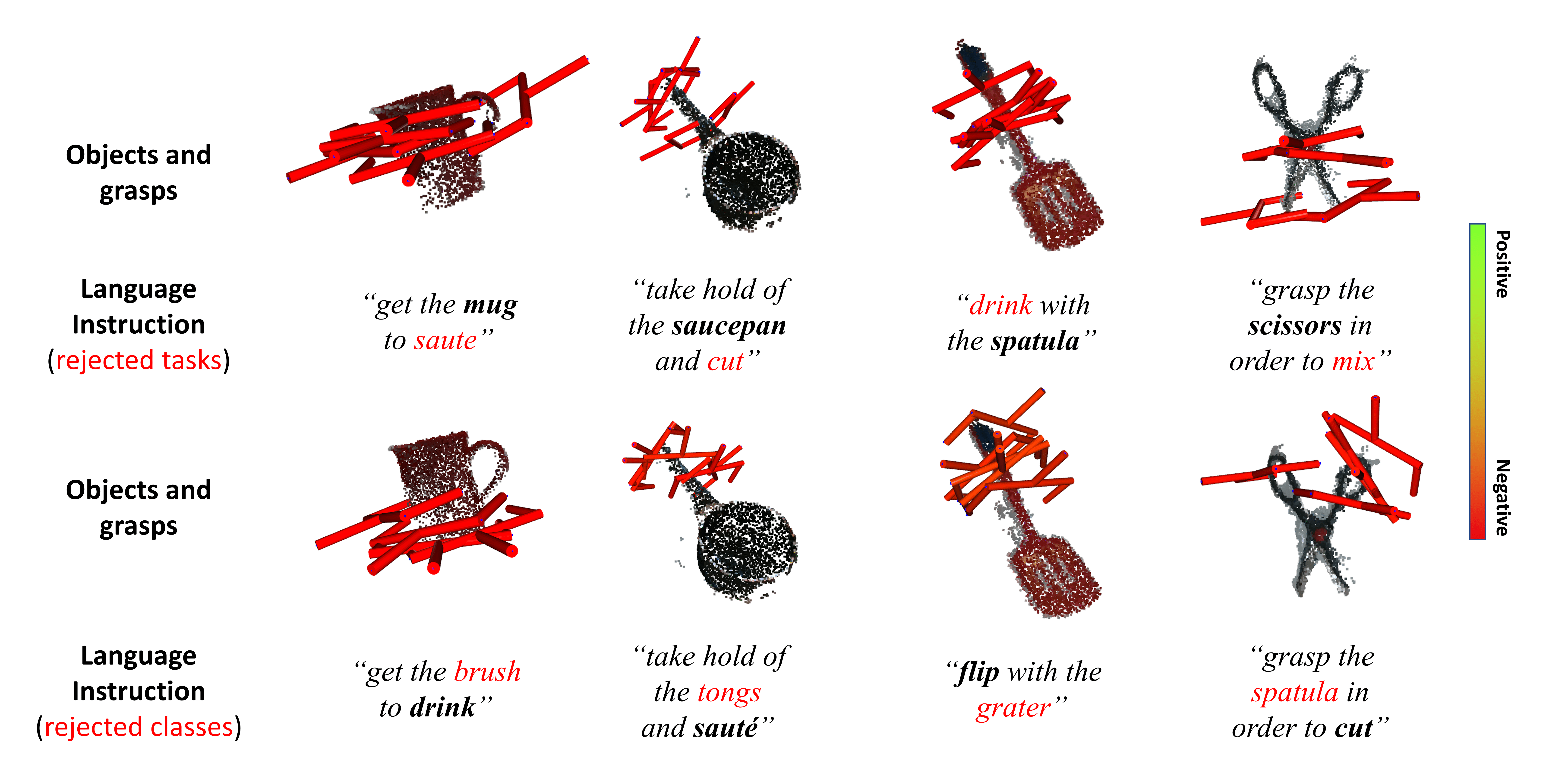}};
  \end{tikzpicture}
    % \vspace*{-0.1in}
  \caption{FoundationGrasp exhibits commonsense reasoning capability. It rejects all grasp candidates under
conditions where the object is incompatible with the
intended task, or there is a discrepancy between the target
and actual object classes.
}
  \label{fig:reject_grasp}
  \vspace*{-0.1in}
\end{figure}

% Please add the following required packages to your document preamble:
% \usepackage{multirow}

\section{Experiments}\label{exp}

\subsection{Results of Perception Experiments}
The quantitative results of perception experiments are visualized in Figure \ref{fig:perception_quantative}. For three held-out settings, TAG establishes a lower performance bound for all methods, achieving approximate mAPs of 50-60\%. The seemingly high mAPs by random guessing is because the distribution of positive and negative samples in the LaViA-TaskGrasp dataset is about even. In the case where the dataset is annotated with more negative samples, mAPs are expected to be much lower. SGN surpasses TAG's performance by over 10\% since it considers the task constraints on object grasping. Moreover, ConceptNet word embeddings capture the semantic similarity between relevant concepts in the embedding space. Consequently, SGN+we achieves a marginal performance boost compared to SGN.

In the held-out instance setting, where no object classes and tasks are held out from the pre-defined KG, GCNGrasp-ow and GCNGrasp-cw are essentially the same, achieving mAP scores of 83.24\%, 82.02\%, and 73.71\%. This is because the semantic KG pre-defines all the object classes, tasks, and their semantic relations within the dataset. However, the performance gap between GCNGrasp-ow and GCNGrasp-cw becomes evident in the other held-out settings. GCNGrasp-ow suffers when generalizing to novel classes and tasks outside the KG. It even falls behind TAG by 1.58\%, 11.03\%, and 19.20\%, and behind SGN by 17.69\%, 24.38\%, and 30.18\% across the three evaluation metrics. In contrast, since GCNGrasp-cw assumes access to all the object classes, tasks, and their ground truth relations in the KG, it performs consistently better. Comparing the two versions of GCNGrasp, the performance degrades by 6.43\%, 4.43\%, and 5.39\% in the held-out class setting, and by 22.95\%, 28.85\%, and 42.94\% in the held-out task setting. The significant degradation highlights the limitation of previous methods. The differences between the two held-out settings suggest that while the pre-trained word embeddings from ConceptNet can effectively capture linguistic relations between object classes, they are less effective at relating task concepts.

With the open-ended knowledge from foundation models, FoundationGrasp and GraspGPT outperform TAG, SGN, and GCNGrasp in both held-out class and task settings. Even though FoundationGrasp does not assume access to all the objects and tasks in the dataset during training as GCNGrasp-cw does, it still outperforms GCNGrasp-cw. Due to the incorporation of geometric knowledge, FoundationGrasp further improves the performance of GraspGPT by approximately 10\%. Overall, FoundationGrasp performs the best in all held-out settings. Regarding running time, it takes an average of 1.88s for FoundationGrasp to evaluate 25 grasp candidates per instance in the LaViA-TaskGrasp on a single RTX 3090. A qualitative comparison between all methods is illustrated in Figure \ref{fig:grasp_qualitative}. Note that even though baselines may assign high confidence scores to their predictions, they often fail to accurately localize the task-oriented regions. FoundationGrasp also exhibits commonsense reasoning capability to some degree. As depicted in Figure \ref{fig:reject_grasp}, FoundationGrasp rejects all grasp candidates (i.e., assigns low confidence scores) under conditions where the object is incompatible with the intended task, or there is a discrepancy between the target and actual object classes.

\begin{table*}[t]
\centering
\renewcommand\arraystretch{2}
\setlength\tabcolsep{5pt}%调列距
\begin{tabular}{cccclcccccccc}
\toprule
\multirow{2}{*}{\textbf{Method}} & \multicolumn{3}{c}{\textbf{Held-out Instance Performance}} & \multirow{2}{*}{Success} & \multicolumn{3}{c}{\textbf{Held-out Class Performance}} & \multirow{2}{*}{Success}    & \multicolumn{3}{c}{\textbf{Held-out Task Performance}} & \multirow{2}{*}{Success}    \\ \cline{2-4} \cline{6-8} \cline{10-12}
                        & Perception        & Planning       & Action       &                          & Perception       & Planning      & Action      &                             & Perception      & Planning      & Action      &                             \\ \midrule
GCNGrasp                & 81/100             & 70/100          & 65/100        & \multicolumn{1}{c}{65.00\%}     & 50/100            & 44/100         & 37/100       &           37.00\%                  & 50/100           & 42/100         & 34/100       &          34.00\%                     \\
GraspGPT              & 83/100             & 69/100          & 64/100        &          64.00\%                  & 80/100           & 69/100        & 65/100      & \multicolumn{1}{l}{65.00\%} & 75/100          & 66/100        & 59/100      & \multicolumn{1}{l}{59.00\%} \\
FoundationGrasp            & 89/100             & 82/100          & 74/100        &          \textbf{74.00\%}                  & 81/100            & 77/100         & 71/100       & \multicolumn{1}{l}{\textbf{71.00\% }}        & 83/100           & 78/100         & 73/100       & \multicolumn{1}{l}{\textbf{73.00\%}}        \\ \bottomrule
\end{tabular}
% \vspace*{0.1in}
\caption{Quantitative results of single-object task-oriented grasping experiments}
\label{tab:grasping_quantitative}
% \vspace*{-0.2in}
\end{table*}

% Please add the following required packages to your document preamble:
% \usepackage{multirow}
\begin{table*}[t]
\centering
\renewcommand\arraystretch{2}
\setlength\tabcolsep{1pt}%调列距
\begin{tabular}{cccccccccccccccc}
\toprule
\multirow{2}{*}{\textbf{Method}} & \multicolumn{4}{c}{\textbf{Light Clutter}}          & \multirow{2}{*}{Success} & \multicolumn{4}{c}{\textbf{Medium Clutter}}         & \multirow{2}{*}{Success} & \multicolumn{4}{c}{\textbf{Heavy Clutter}}          & \multirow{2}{*}{Success} \\ \cline{2-5} \cline{7-10} \cline{12-15}
                                 & Grounding & Perception & Planning & Action &                          & Grounding & Perception & Planning & Action &                          & Grounding & Perception & Planning & Action &                          \\ \hline
FoundationGrasp                  & 14/15     & 13/15      & 12/15    & 12/15  & 80.00\%                  & 13/15     & 11/15      & 11/15    & 11/15  & 73.33\%                  & 10/15     & 9/15       & 8/15     & 8/15   & 53.33\%                  \\ \bottomrule
\end{tabular}
\caption{Quantitative results of multi-object task-oriented grasping experiments}
\label{tab:mo_grasping_quantitative}
\end{table*}

\begin{figure}[t]
  \centering
  \vspace*{-0.1in}
  \begin{tikzpicture}[inner sep = 0pt, outer sep = 0pt]
    \node[anchor=south west] (fnC) at (0in,0in)
      {\includegraphics[height=2.0in,clip=true,trim=0in 0in 0in 0in]{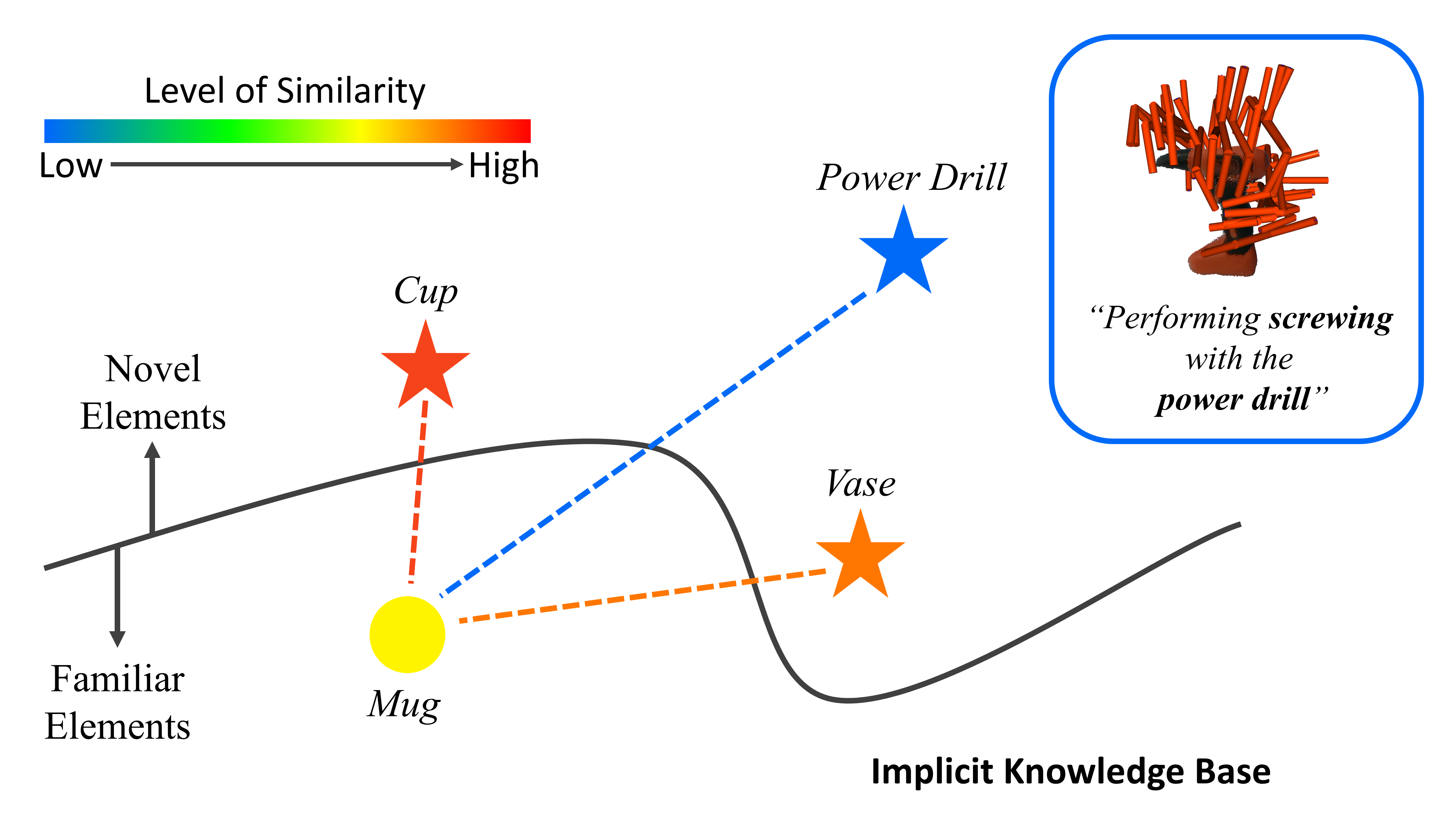}};
  \end{tikzpicture}
    \vspace*{-0.3in}
  \caption{Since the novel object class ``\textit{Power Drill}" shares minimal similarity to familiar elements in the implicit knowledge base, FoundationGrasp makes false predictions by rejecting all valid candidates. The colors of novel elements denote the level of similarity to the ``\textit{Mug}".}
  \label{fig:false_neg_concept}
  \vspace*{-0.2in}
\end{figure}

\begin{figure*}[t]
  \centering
  % \vspace*{-0.2in}
  \begin{tikzpicture}[inner sep = 0pt, outer sep = 0pt]
    \node[anchor=south west] (fnC) at (0in,0in)
      {\includegraphics[height=2.2in,clip=true,trim=0in 0in 0in 0in]{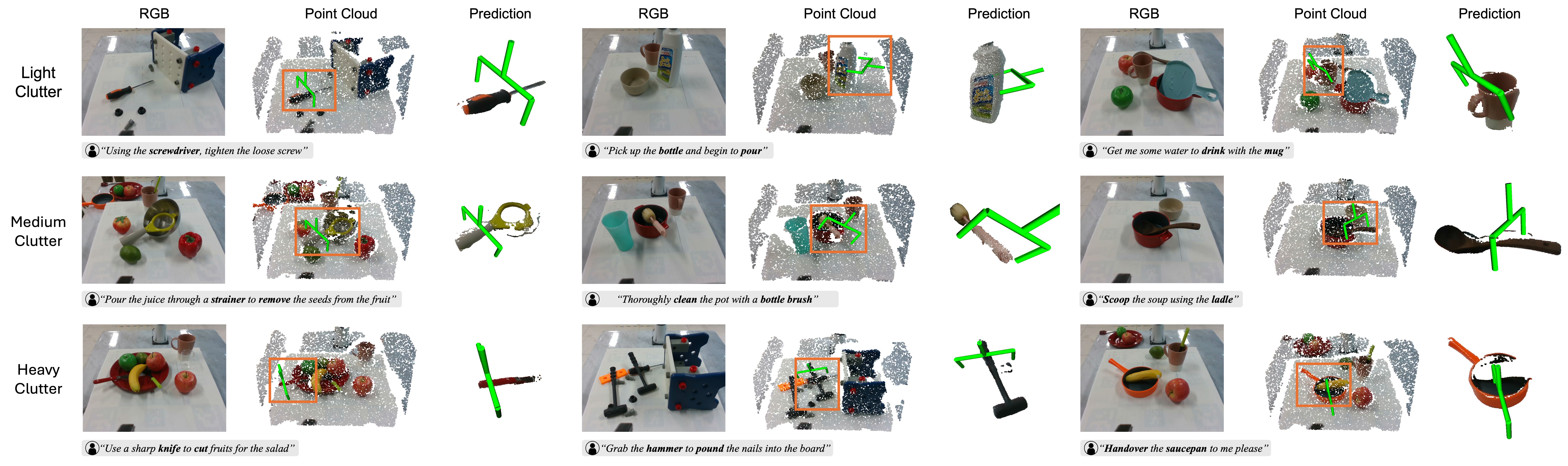}};
  \end{tikzpicture}
    % \vspace*{-0.28in}
        \caption{Qualitative results of multi-object task-oriented grasping experiments}
  \label{fig:mo_tog_qualitative}
  \vspace*{-0.1in}
\end{figure*}

\subsection{Results of Real-Robot Experiments}

\textit{Single-Object Task-Oriented Grasping Experiments} \ The results of single-object task-oriented grasping experiments are summarized in Table \ref{tab:grasping_quantitative}. Despite capturing object point clouds from a single viewpoint, FoundationGrasp consistently achieves a perception success rate exceeding 80\% across all settings. Similar to earlier observations in perception experiments, GCNGrasp suffers when generalizing to novel classes and tasks, while GraspGPT and FoundationGrasp perform consistently better. The improvement observed with FoundationGrasp over GraspGPT indicates the necessity of incorporating both semantic and geometric knowledge. The running time of FoundationGrasp during the perception stage is illustrated in Table \ref{tab:runtime}. Compared to perception experiments where object and task knowledge in the LaViA-TaskGrasp dataset is pre-generated and pre-processed, it takes an extra 90.40s to prompt foundation models and process the knowledge on the fly in real-robot experiments. FoundationGrasp experiences a significant performance drop from the perception to the action stage. Three typical failure modes are identified: (1) unstable grasp candidates generated by the task-agnostic grasp sampler, (2) motion planning failures, and (3) false rejection of valid candidates. Regarding the third failure mode, the training stage of FoundationGrasp can be conceptualized as constructing an implicit knowledge base in the latent space with a set of objects and tasks. During the inference stage, FoundationGrasp aims to build connections between existing elements in the knowledge base and novel ones outside. However, if novel elements are substantially divergent from the knowledge base, FoundationGrasp tends to make wrong predictions. As illustrated in Figure \ref{fig:false_neg_concept}, when instructing the robot to screw with a power drill, the object class \textit{power drill} shares minimal semantic or geometric similarity to existing elements in the knowledge base, leading to the rejection of all valid candidates. This issue can potentially be resolved by constructing a base element set that is representative enough yet compact. 

\begin{table}[t]
\centering
\renewcommand\arraystretch{2}
\setlength\tabcolsep{4pt}%调列距
\begin{tabular}{cccccc}
\toprule
\multirow{2}{*}{Phase} & \multicolumn{4}{c}{Perception Stage}             & \multirow{2}{*}{Total} \\ \cline{2-5}
                       & Segmentation & Sampling & Prompting & Evaluation &                        \\ \hline
Time(s)                & 7.51         & 4.32     & 90.40     & 1.88       & 104.11                 \\ \bottomrule
\end{tabular}
\caption{Running time of FoundationGrasp in real-robot experiments}
\label{tab:runtime}
  \vspace*{-0.3in}
\end{table}

\textit{Multi-Object Task-Oriented Grasping Experiments} \ To further assess the limits of FoundationGrasp in more complex and realistic environments, we conduct multi-object TOG experiments with three levels of difficulty. The results are presented in Table \ref{tab:mo_grasping_quantitative}. The success rates remain high (above 70\%) in both light and medium clutter settings. This can be attributed to two factors: (1) GroundingDINO+SAM’s ability to accurately detect and segment target objects in less cluttered scenes, and (2) FoundationGrasp’s enhanced geometric understanding, enabling it to interpret the geometric structures of objects and tasks even with partial observations and light occlusions. However, the success rate in heavy clutter drops below 60\%. This decrease is expected since heavy occlusions and contacts lead to limited visibility of target objects and complicate the search for collision-free paths to reach them. Nevertheless, the small gap between the grounding and action stages suggests that FoundationGrasp remains effective as long as there is adequate visual observation.

% explain the third failure mode, connection training and testing datatbase, ig there is no similarity, not able to generalize, meta-database, meta object class and task selection are important

\begin{figure*}[t]
  \centering
  \vspace*{-0.2in}
  \begin{tikzpicture}[inner sep = 0pt, outer sep = 0pt]
    \node[anchor=south west] (fnC) at (0in,0in)
      {\includegraphics[height=2.6in,clip=true,trim=0in 0in 0in 0in]{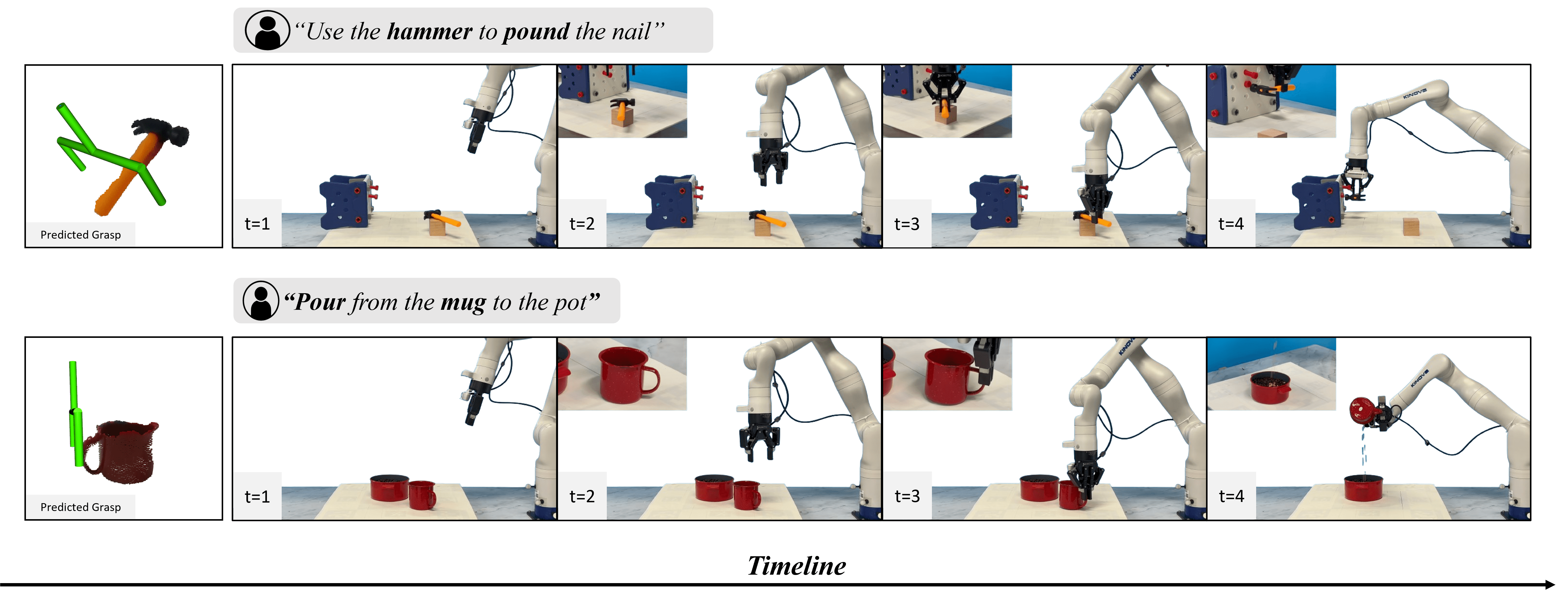}};
  \end{tikzpicture}
    % \vspace*{-0.28in}
        \caption{Qualitative results of task-oriented manipulation experiments. The robot predicts task-oriented grasps with FoundationGrasp and generates manipulation trajectories by merging sequential motion primitives.}

  \label{fig:mani_qualitative}
  \vspace*{-0.1in}
\end{figure*}

\textit{Task-Oriented Manipulation Experiments} \ This evaluation aims to prove that task-oriented grasping facilitates downstream manipulation tasks. Therefore, baselines are not evaluated here. As reported in Table \ref{tab:mani_quantitative}, FoundationGrasp achieves a high grasping success rate of approximately 80\% on nine manipulation tasks. Specifically, in the Bottle Opening task, since the bottle cap is flat and thin, the robot attempts to grasp at the cap-neck intersection in some cases, resulting in 5 failures out of 10 trials. In terms of manipulation, despite simplifying the manipulation motions into motion primitives guided by rule-based heuristics, FoundationGrasp performs reasonably well, with an average success rate of 66.67\%. Two examples of task-oriented manipulation experiments are presented in Figure \ref{fig:mani_qualitative}. The performance gap between the grasping and manipulation stages is primarily due to two factors: (1) limited ability to model the relative pose between the tool and the receptacle objects; (2) absence of force feedback in tasks requiring precise contact control, such as Mug Cleaning and Ketchup Squeezing. Future work will consider extending FoundationGrasp to jointly optimize grasp poses and manipulation trajectories and incorporate force feedback.

In summary, there are three primary challenges encountered in real-robot experiments:
\begin{itemize}
    \item Limited Generalization: A significant challenge is the limited ability to generalize to novel objects and tasks unseen during training. FoundationGrasp addresses this by leveraging open-ended knowledge from foundation models, thereby enhancing the generalization capabilities beyond those of previous TOG methods.
    \item Occlusion: Another challenge arises from occlusions. If the target object or functional part is significantly occluded, the robot may struggle to generate valid grasp candidates or may produce false evaluations.
    \item Planning: The third challenge involves planning. Since FoundationGrasp only focuses on generating TOG poses for the end-effector, the off-the-shelf planner is utilized to move the gripper to the target pose, during which planning failures might occur.
\end{itemize}

% (1) The first challenge is limited generalization to novel novel objects and tasks unseen during training. By leveraging the open-ended knowledge from foundation models, FoundationGrasp significantly improves the generalization capabilities of previous TOG methods. (2) The second challenge comes from perception. The quality of the perception heavily influences the grasping performance. For example, under our single-view setting, if the mug handle is invisible to the robot, the robot cannot generate grasps on the point cloud of the handle for drinking. Reconstruction from multiple views might alleviate this problem. However, this is not the focus of this paper. (3) The third challenge is planning. Since FoundationGrasp only focuses on generating TOG poses of the end-effector, the authors use the off-the-shelf planner (e.g., RRT-Connect) to move the gripper to the target pose, during which planning failures might occur.  

% \cite{pan2023tax} 

% More qualitative results of real-robot experiments can be found in the appendix.

\begin{table}[t]
\renewcommand\arraystretch{1.8}
\setlength\tabcolsep{6pt}%调列距
\begin{tabular}{lccc}
\toprule
\multicolumn{1}{c}{\multirow{2}{*}{\textbf{Household Task}}} & \multirow{2}{*}{\textbf{Household Object}} & \multicolumn{2}{c}{Success} \\ \cline{3-4} 
\multicolumn{1}{c}{}                                   &                                   & Grasping   & Manipulation   \\ \hline
Dish Washing                                           & Plate, Scrubber                   & 8/10       & 7/10           \\
Bean Scooping                                          & Spoon, Pot                        & 9/10       & 9/10           \\
Nail Pounding                                          & Nail, Hammer                      & 9/10       & 6/10           \\
Knife Handover                                         & Knife                             & 8/10       & 8/10           \\
Bean Pouring                                           & Mug, Bowl                         & 8/10       & 6/10           \\
Bottle Opening                                         & Bottle                            & 5/10       & 5/10           \\
Ingredients Mixing                                     & Spoon, Pot                        & 9/10       & 7/10           \\
Mug Cleaning                                           & Mug, Sponge Brush                 & 7/10       & 6/10           \\
Ketchup Squeezing                                      & Squeeze Bottle, Plate             & 8/10       & 6/10           \\ \hline
\textbf{Total}                        & \multicolumn{1}{l}{}              & 78.89\%    & 66.67\%        \\ \bottomrule
\end{tabular}
% \vspace*{0.1in}
\caption{Quantitative results of task-oriented manipulation experiments}
\label{tab:mani_quantitative}
\vspace*{-0.1in}
\end{table}

% \vspace*{-0.2in}
\subsection{Ablation Study}

% (1) sem-go, (2) obj-task, (3) fusion, (4) VLMS, (5) LLMs, (6) lan coders 
To gain further insights into the design choices of FoundationGrasp, we conduct a series of ablation studies to answer the following questions:

\begin{itemize}
\item How is the performance of FoundationGrasp affected by the semantic and geometric branches?
\item How do object and task knowledge contribute to the generalization to novel elements outside the training set?
\item How do different fusion strategies in the task-oriented grasp evaluator affect the performance of FoundationGrasp?
\item How does the choice of a VLM in the geometric knowledge encoder impact the performance of FoundationGrasp?
\item How does the choice of an LLM in the semantic knowledge encoder impact the performance of FoundationGrasp?
\item How does the choice of an LLM in language description generation impact the performance of FoundationGrasp?
\end{itemize}

\begin{figure}[t]
  \centering
  \vspace*{-0.2in}
  \begin{tikzpicture}[inner sep = 0pt, outer sep = 0pt]
    \node[anchor=south west] (fnC) at (0in,0in)
      {\includegraphics[height=4.5in,clip=true,trim=0.3in 0in 0in 0in]{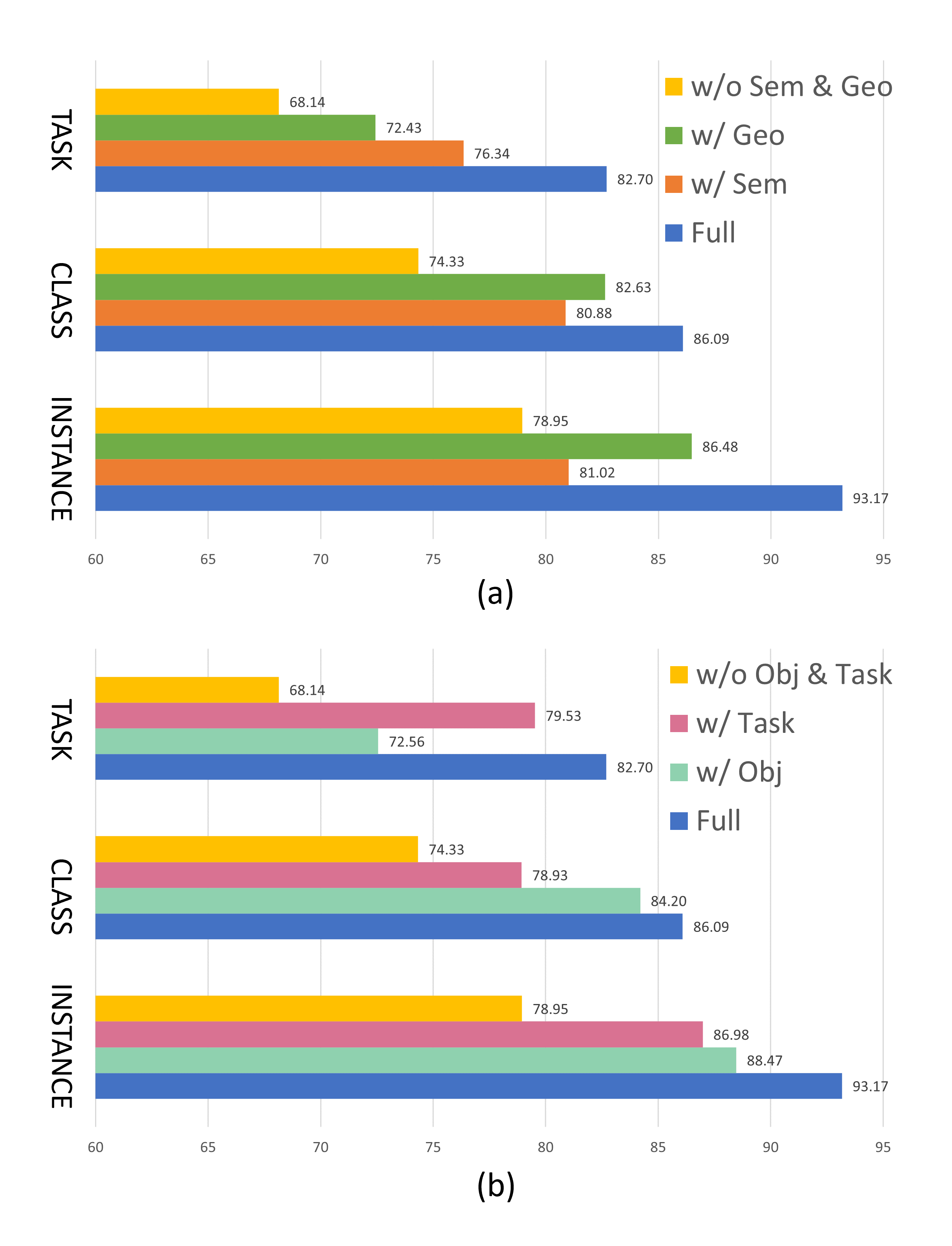}};
  \end{tikzpicture}
    \vspace*{-0.1in}
  \caption{(a) Ablation study on semantic and geometric branches. (b) Ablation study on object and task knowledge.}
  \label{fig:ablation_study}
  \vspace*{-0.2in}
\end{figure}

% sem and geo knowledge
\textit{Ablation on Semantic and Geometric Branches} \ In this study, we compare FoundationGrasp to three ablated versions: (1) with the semantic branch only (i.e., w/ Sem); (2) with the geometric branch only (i.e., w/ Geo); (3) the vanilla model (i.e., w/o Sem \& Geo). The results are presented in Figure \ref{fig:ablation_study} (a). Here, we only show mAPs of the corresponding held-out settings. For all held-out settings, the full model achieves the best performance, while the vanilla model performs the worst. This demonstrates the effectiveness of the semantic-geometric branch architecture. Specifically, for the held-out instance setting, where objects have limited variations in terms of geometric and visual features, geometric knowledge is essential for unifying these variances. The relations between tasks are easier to capture when tasks are abstracted as linguistic concepts, making semantic knowledge more critical in relating different tasks in the held-out task setting. To conclude, both semantic and geometric branches play equally important roles when generalizing to novel object classes.

% object and class knowledge
\textit{Ablation  on Object and Task Knowledge} \ In this study, we compare FoundationGrasp to three ablated versions: (1) with object knowledge only (i.e., w/ Obj); (2) with task knowledge only (i.e., w/ Task); (3) the vanilla model (i.e., w/o Obj \& Task). The results are presented in Figure \ref{fig:ablation_study} (b). It is clear from the held-out class and task settings that object and task knowledge are crucial for generalizing to novel object classes and tasks, respectively. This finding aligns with our intuition. In the held-out class setting, the full model only outperforms w/ Obj by 1.89\%. For the held-out instance setting, w/ Obj and w/ Task achieve similar performances (88.47\% and 86.98\%). Overall, incorporating both object and task knowledge leads to superior performance compared to all ablated versions.

Since each of the following ablation studies shares similar conclusions across the three held-out settings, we only report the results of the held-out class setting for simplicity.

\textit{Ablation on Fusion Strategy} \  In this study, we investigate how different fusion strategies impact the performance of FoundationGrasp. Specifically, we compare the original cross-attention-based fusion (denoted as Cross-Attn) with the following alternative methods: (1) double attention \cite{chen20182} (denoted as Double-Attn), which replaces the cross attention in each transformer block with two consecutive attention blocks for feature gathering and distribution; (2) spatial-channel attention \cite{chen2017sca} (denoted as SC-Attn), which uses spatial and channel attention in parallel; (3) feature-wise linear modulation \cite{perez2018film} (denoted as FiLM), which modifies features via a learnable affine transformation based on conditioning information; and (4) concatenation (denoted as Concat), which concatenates all input features channel-wise in each transformer block without any learnable parameters. As shown in Table \ref{tab: ablation_fusion}, the simple concatenation fusion achieves the worst performance, falling behind Cross-Attn by 13.51\%, 13.75\%, and 10.42\% across the three evaluation metrics. Among the attention-based fusion strategies, SC-Attn demonstrates a slight advantage over Cross-Attn, with improvements of 0.75\% in Instance mAP and 0.85\% in Class mAP. Despite these marginal gains, we use Cross-Attn due to its balance between performance and computational cost. 

\begin{table}[h]
\centering
\renewcommand\arraystretch{1.8}
\setlength\tabcolsep{6pt}%调列距
\begin{tabular}{cccc}
\toprule
\multirow{2}{*}{\textbf{Method}} & \multicolumn{3}{c}{\textbf{Held-out Class Performance} (\%)} \\ \cline{2-4} 
                                 & Instance mAP        & Class mAP        & Task mAP       \\ \hline
Cross-Attn                       & 88.74               & 86.09            & 84.33          \\
Double-Attn                      & 85.43               & 83.43           & 80.78          \\
SC-Attn                          & 89.49               & 86.94            & 84.01          \\
FiLM                             & 83.87               & 80.54            & 81.29          \\
Concat                           & 75.23               & 72.34            & 73.91          \\ \bottomrule
\end{tabular}
\caption{Ablation study on fusion strategy}
\label{tab: ablation_fusion}
  % \vspace*{-0.2in}
\end{table}
\textit{Ablation on LLMs in Semantic Knowledge Encoder} \ In this study, we equip FoundationGrasp with four different LLMs, including BERT, RoBERTa\cite{liu2019roberta}, LLaMA 3\cite{touvron2023llama}, and T5\cite{raffel2020exploring}, in the semantic knowledge encoder and compare their performances. The results are detailed in Table \ref{tab:ablation_lan_encoder}. Overall, all LLMs perform comparably, with LLaMA 3 showing a slight edge with mAPs of 89.63\%, 87.53\%, and 86.48\%. The consistent performance across the LLMs is due to their extensive pre-training on large text corpora, which endows them with a robust understanding of commonsense semantic knowledge. Despite the similarities, we select Google pre-trained BERT as the semantic knowledge encoder in FoundationGrasp, balancing effective knowledge understanding with computational efficiency.

\begin{table}[t]
\centering
\renewcommand\arraystretch{1.8}
\setlength\tabcolsep{6pt}%调列距
\begin{tabular}{cccc}
\toprule
\multirow{2}{*}{\textbf{Method}} & \multicolumn{3}{c}{\textbf{Held-out Class Performance} (\%)} \\ \cline{2-4} 
                                 & Instance mAP        & Class mAP        & Task mAP       \\ \hline
BERT                             & 88.74               & 86.09            & 84.33          \\
RoBERTa                          & 87.54               & 86.45            & 84.09          \\
LLaMA 3                          & 89.63               & 87.53            & 86.48          \\
T5                               & 88.73               & 87.90             & 85.34          \\ \bottomrule
\end{tabular}
\caption{Ablation study on LLMs in the semantic knowledge encoder}
\label{tab:ablation_lan_encoder}
  \vspace*{-0.2in}
\end{table}

\textit{Ablation on VLMs in Geometric Knowledge Encoder} \ In this study, we explore the impact of different VLMs by substituting the CLIP model used in the geometric knowledge encoder with three alternatives: (1) BLIP-2 \cite{li2023blip}, noted for its strong performance in multi-modal tasks like Visual Question Answering; (2) LLaVa \cite{liu2024visual}, which combines a CLIP vision encoder with a LLaMA language encoder; and (3) CLIP-V, which removes the CLIP language encoder. The results, summarized in Table \ref{tab: ablation_vlm}, show that all three VLMs perform comparably, with CLIP achieving the highest performance in Class mAP (86.09\%) and Task mAP (84.33\%). Notably, CLIP-V, which relies solely on visual inputs, underperforms relative to CLIP by 5.21\%, 4.26\%, and 3.35\% across the evaluation metrics. This finding emphasizes the significance of multi-modal knowledge for effective geometric understanding, as supported by previous studies \cite{kerr2023lerf, qin2024langsplat}.

\begin{table}[h]
\centering
\renewcommand\arraystretch{1.8}
\setlength\tabcolsep{6pt}%调列距
\begin{tabular}{cccc}
\toprule
\multirow{2}{*}{\textbf{Method}} & \multicolumn{3}{c}{\textbf{Held-out Class Performance} (\%)} \\ \cline{2-4} 
                                 & Instance mAP        & Class mAP        & Task mAP       \\ \hline
CLIP                             & 88.74               & 86.09            & 84.33          \\
BLIP-2                           & 86.49               & 84.92            & 82.09          \\
LLaVa                            & 88.98               & 85.43            & 84.01          \\
CLIP-V                           & 83.53               & 81.83            & 80.98          \\ \bottomrule
\end{tabular}
\caption{Ablation study on VLMs in the geometric knowledge encoder}
\label{tab: ablation_vlm}
  % \vspace*{-0.2in}
\end{table}

\textit{Ablation on LLMs in Language Description Generation} \ In this study, we evaluate the performance of FoundationGrasp using two different LLMs for language description generation: GPT-4 and GPT-3.5. As expected, the more advanced GPT-4 slightly outperforms GPT-3.5 across the three evaluation metrics by 0.76\%, 1.7\%, and 1.24\%, respectively. However, the performance differences between the two models are relatively minor.

\begin{table}[h]
\centering
\renewcommand\arraystretch{1.8}
\setlength\tabcolsep{6pt}%调列距
\begin{tabular}{cccc}
\toprule
\multirow{2}{*}{\textbf{Method}} & \multicolumn{3}{c}{\textbf{Held-out Class Performance} (\%)} \\ \cline{2-4} 
                                 & Instance mAP        & Class mAP        & Task mAP       \\ \midrule
GPT-4                            & 88.74               & 86.09            & 84.33          \\
GPT-3.5                          & 87.98               & 84.39            & 83.09          \\ \bottomrule
\end{tabular}
\caption{Ablation study on LLMs for language description generation}
  \vspace*{-0.2in}
\end{table}

To summarize, the key insights from the ablation studies on foundation models are two-fold:

\begin{itemize}
    \item FoundationGrasp demonstrates robust performance across various VLMs and LLMs, with only minor performance differences between different foundation models.
    \item The modular design of FoundationGrasp allows for the seamless integration of any state-of-the-art foundation models. This adaptability ensures that as foundation models evolve, the performance of FoundationGrasp will also improve accordingly.
\end{itemize}

\subsection{User Study}
To deploy FoundationGrasp on household robots operating in everyday life, it is also necessary to study its practical applicability from the user perspective (e.g., whether the behavior of FoundationGrasp aligns with user needs, preferences, safety, and comfort). Therefore, we conduct a user study with 10 participants on 10 test objects selected from the real-robot experiments benchmark. Once all objects are tested, participants are invited to fill out a questionnaire with a Likert scale. Participants are also encouraged to share their experiences or describe any problems they encounter during the user study. 

% . A set of 10 test objects is selected from the benchmark used in real-robot experiments. The robot is allowed to make five attempts at most for each object.

% For each trial, an object is placed by the participant in the robot workspace with a random position and orientation. A free-form language instruction is then issued by the participant. 

\begin{figure}[t]
  \centering
  % \vspace*{-0.3in}
  \begin{tikzpicture}[inner sep = 0pt, outer sep = 0pt]
    \node[anchor=south west] (fnC) at (0in,0in)
      {\includegraphics[height=1.4in,clip=true,trim=0.15in 0in 0in 0in]{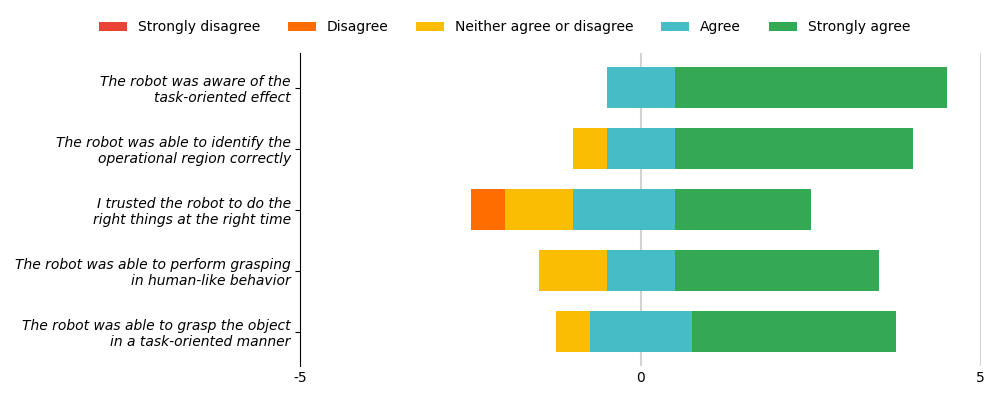}};
  \end{tikzpicture}
    \vspace*{-0.2in}
  \caption{Participants’ agreement with each statement in the user study questionnaire. Horizontal
axis denotes levels of agreement w.r.t. each statement, where Strongly Agree-(5), Agree-(2.5), Neither Agree or Disagree-(0), Disagree-(-2.5), and Strongly Disagree-(5).}
  \label{fig:user_study}
  \vspace*{-0.2in}
\end{figure}

Table \ref{tab:user_study} reports the objective metrics of the user study. On average, the robot can successfully grasp each object with less than two attempts. Figure \ref{fig:user_study} presents the subjective responses of participants to the questionnaire. Average ratings in all dimensions are above 3 (i.e., Agree or Strongly Agree). Key findings from the user study include: (1) Most participants agree that the robot is aware of the task-oriented effect and can grasp objects in a task-oriented manner; (2) Some comment that FoundationGrasp exhibits high repeatability, generating grasp poses around the same point for the same object and task with different expressions; (3) Participants find that FoundationGrasp is robust to arbitrary object poses, attributing this to the data augmentation performed during training. Regarding safety, one participant expresses concerns about potential collisions with users, resulting in a disagreement on the third statement. Overall, participants suggest the system could be improved in the following ways: (1) The system should consider the state of the object. For example, the generated grasps should be far away from the body of the pot if the pot is hot. (2) The system should be able to understand language instructions directly specifying the part (e.g., ``\textit{grasp the handle of the pot}").

% Particular objects, such as a mustard bottle, cereal box, and spoon, take more trials for a successful grasp. It is either due to noisy point clouds caused by the reflective material or motion planning failure caused by the relatively large size of the object. 

\begin{table*}[t]
\centering
\setlength\tabcolsep{4pt}%调列距
\renewcommand\arraystretch{1.8}
\begin{tabular}{lcccccccccc}
\toprule
\multicolumn{1}{c}{} & Hammer   & Brush    & Scissors & Cereal Box & Saucepan & Mustard Bottle & Spatula  & Bottle Brush & Spoon    & Screwdriver \\ \hline
Number of Attempts   & 1.3$\pm$0.15 & 1.2$\pm$0.20 & 1.4$\pm$0.22 & 1.5$\pm$0.27   & 1.3$\pm$0.15 & 1.7$\pm$0.26       & 1.1$\pm$0.10 & 1.0          & 1.5$\pm$0.40 & 1.4$\pm$0.40    \\
Success Rate         & 70\%       & 90\%       & 70\%       & 70\%         & 70\%       & 50\%             & 90\%       & 100\%          & 80\%       & 80\%          \\ \bottomrule
\end{tabular}
  % \vspace{0.1in}
\caption{Objective evaluation of user study}
\label{tab:user_study}
  \vspace{-0.2in}
\end{table*}

\section{Limitations and Future Work}\label{discuss}

% add outcomes and compare them with existing works
The results of perception experiments verify the claim that FoundationGrasp outperforms existing TOG methods when generalizing to novel object instances, object classes, and tasks out of the training set. Meanwhile, real-robot experiments have also demonstrated the effectiveness of FoundationGrasp in real-world applications. Despite compelling results, we observe typical failure modes or limitations during experiments. This section discusses these limitations from data, algorithm, and hardware perspectives and provides insights into future work. 

% The data annotation phase requires Amazon Mechanical Turk (AMT) annotators to crowdsource labels for the 250K task-agnostic grasps, which is labor-intensive. 

% text-to-image (e.g., DALL-E\cite{ramesh2021zero}) and text-to-video

\textit{Grasp Data Acquisition} \ While FoundationGrasp achieves superior generalization performance compared to previous works, the data collection phase still requires labor-intensive manual annotation. Additionally, some annotations do not align well with how humans perform task-oriented grasping, which may further hurt performance. This issue is reflected by the neutral attitude of several participants towards the statement “The robot was able to perform grasping in human-like behavior” in the user study. One potential solution is to learn from human demonstrations. Previous works have shown that human demonstrations (e.g., images or videos) encode constraints and semantics of human behaviors for robot skill learning. Learning with general-purpose simulators of the physical world is also an exciting direction. Recent advances in AIGC (e.g., Sora \cite{sora2024}) open up the potential for generating synthetic demonstration data without a human in the loop.

\textit{Sampling-and-Evaluation Pipeline} \ FoundationGrasp currently follows a two-stage, sampling-and-evaluation pipeline. Using the off-the-shelf task-agnostic grasp sampler reduces the complexity of constructing a TOG pipeline. However, such a design also leads to two problems. First, we assume all grasp candidates from the sampling phase are stable and treat them equally in the evaluation phase. However, we observe that FoundationGrasp tends to bias marginal grasp poses, such as grasps near the end of the handle. Although functionally correct, these grasp poses are not configurationally robust to perturbations, such as unexpected contact or calibration error. This would lead to potential grasping failure. Second, the first stage samples grasps uniformly over the point cloud. However, robots would only interact with specific regions of an object. Exhaustively evaluating every candidate grasp significantly increases the computational cost of FoundationGrasp and hinders real-time prediction. Future work plans to merge task-agnostic sampling with task-oriented evaluation into a unified framework.  

\textit{Robot Gripper Generalization} \ Regarding hardware setup, we currently only consider the parallel jaw gripper. Therefore, FoundationGrasp cannot be directly transferred to other types of grippers, such as three-finger grippers and five-finger dexterous hands. Although such a methodology can be generally applied to training any gripper-dependent TOG model, it is still far from a universal TOG method in terms of both software and hardware. Ideally, the algorithm would take both object geometry and gripper attributes as inputs and output the grasp poses or contact points. The long-term goal is to develop a universal TOG method that can be readily applied to any household robot with any type of gripper. 

% \textit{Joint Grasp and Motion Optimization} 

% sample and evaluate -> candidate grasp pose end-to-end diffusion model

% grasp pose and genetation and motion planning

\section{Conclusion}\label{conclusion}
In this paper, we present FoundationGrasp, a foundation model-based TOG framework. Compared to previous works, which limit the prior knowledge to a closed-set scope, FoundationGrasp leverages the open-ended knowledge from foundation models to learn generalizable TOG skills. Evaluation on the LaViA-TaskGrasp dataset demonstrates the superiority of FoundationGrasp over existing TOG methods when generalizing to novel object instances, object classes, and tasks. Furthermore, the effectiveness of FoundationGrasp is validated in task-oriented grasping and manipulation experiments on a 7-DoF robotic arm. Lastly, limitations and potential solutions are thoroughly discussed to provide venues for future work.

\bibliographystyle{IEEEtran}
 % \balance
\bibliography{root}

\newpage

% \section{Biography Section}
% If you have an EPS/PDF photo (graphicx package needed), extra braces are
%  needed around the contents of the optional argument to biography to prevent
%  the LaTeX parser from getting confused when it sees the complicated
%  $\backslash${\tt{includegraphics}} command within an optional argument. (You can create
%  your own custom macro containing the $\backslash${\tt{includegraphics}} command to make things
%  simpler here.)
 
% \vspace{11pt}

\vspace{-45pt}

% \bf{If you include a photo:}\vspace{-33pt}
\begin{IEEEbiography}[{\includegraphics[width=1in,height=1.25in,clip,keepaspectratio]{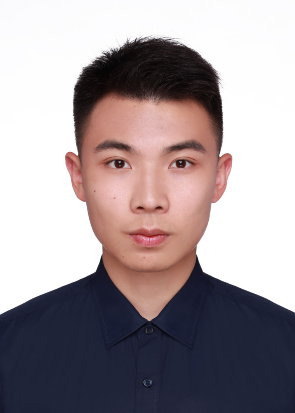}}]{Chao Tang} received his B.S. degree from Nanjing Tech University in 2018 and his M.S. degree from Georgia Institute of Technology in 2020. He is currently pursuing his Ph.D. degree in Robotics at the Southern University of Science and Technology. His research interests involve robotic manipulation and grasping, mobile manipulation, human-robot interaction, and semantic reasoning for robots. 
\end{IEEEbiography}

\vspace{-45pt}

\begin{IEEEbiography}[{\includegraphics[width=1in,height=1.25in,clip,keepaspectratio]{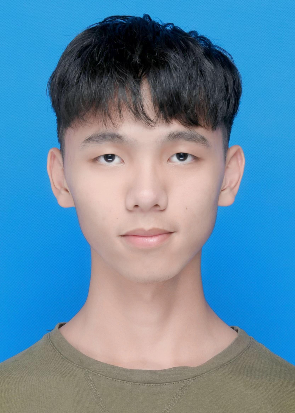}}]{Dehao Huang} received a B.E. degree in information engineering from the South China Normal University, Guangzhou, China, in 2022. He is working toward an M.Sc. degree with the Department of Electrical and Electronic Engineering at the Southern University of Science and Technology, Shenzhen, China. His research interests include scene understanding in robotics and SLAM.

\end{IEEEbiography}

\vspace{-45pt}

\begin{IEEEbiography}[{\includegraphics[width=1in,height=1.25in,clip,keepaspectratio]{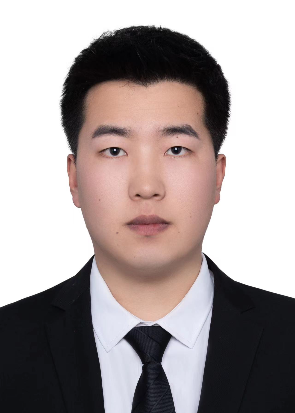}}]{Wenlong Dong} received a B.S. degree from the School of Information Science and Technology, Dalian Maritime University, and is currently pursuing a Master's degree at the Southern University of Science and Technology. His research interests include intelligent robotic manipulation and robot navigation.
\end{IEEEbiography}

\vspace{-45pt}

\begin{IEEEbiography}[{\includegraphics[width=1in,height=1.25in,clip,keepaspectratio]{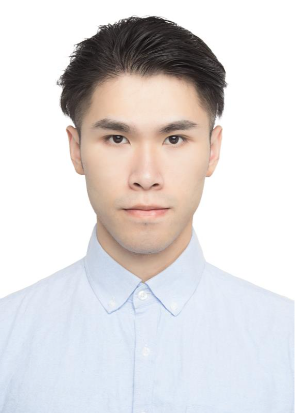}}]{Ruinian Xu} is currently a PhD candidate in Electrical and Computer Engineering at the Georgia Institute of Technology. His current research interests include computational vision, natural language processing, multimodal learning, and graph neural networks for robotics.
\end{IEEEbiography}

\vspace{-45pt}

\begin{IEEEbiography}[{\includegraphics[width=1in,height=1.25in,clip,keepaspectratio]{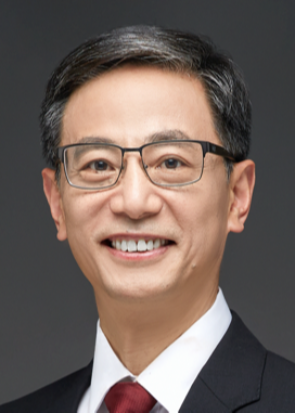}}]{Hong Zhang}
received his PhD in Electrical Engineering from Purdue University in 1986. He was a Professor in the Department of Computing Science, University of Alberta, Canada for over 30 years before he joined Southern University of Science and Technology (SUSTech), China, in 2020, where he is currently a Chair Professor.  Dr. Zhang served as the Editor-in-Chief of IROS Conference Paper Review Board (2020-2022) and is currently a member of the IEEE Robotics and Automation Society Administrative Committee (2023-25). He is a Fellow of IEEE and a Fellow of the Canadian Academy of Engineering.  His research interests include robotics, computer vision, and image processing.
\end{IEEEbiography}
% \vspace{11pt}

% \bf{If you will not include a photo:}\vspace{-33pt}
% \begin{IEEEbiographynophoto}{John Doe}
% Use $\backslash${\tt{begin\{IEEEbiographynophoto\}}} and the author name as the argument followed by the biography text.
% \end{IEEEbiographynophoto}

% \vfill

\end{document}